\documentclass[final]{cvpr}

\usepackage{times}
\usepackage{epsfig}
\usepackage{graphicx}
\usepackage{amsmath}
\usepackage{amssymb}
\usepackage{makecell}
\usepackage{array}
\usepackage{float}
\usepackage{placeins}
\usepackage[skip=2pt]{caption}

\usepackage[pagebackref=true,breaklinks=true,colorlinks,bookmarks=false]{hyperref}

\def\cvprPaperID{****} % *** Enter the CVPR Paper ID here
\def\confYear{CVPR 2021}

\begin{document}

%%%%%%%%% TITLE
\title{TransCAM: Transformer Attention-based CAM Refinement for Weakly Supervised Semantic Segmentation}

\author{Ruiwen Li\textsuperscript{\rm 1}, Zheda Mai\textsuperscript{\rm 2}, Chiheb Trabelsi\textsuperscript{\rm 1}, Zhibo Zhang\textsuperscript{\rm 1}, Jongseong Jang\textsuperscript{\rm 3}, Scott Sanner\textsuperscript{\rm 1}\\
\textsuperscript{\rm 1}University of Toronto, \textsuperscript{\rm 2}Optimy AI, \textsuperscript{\rm 3}LG AI Research\\
% Institution1 address\\
{\tt\small \{ruiwen.li, zheda.mai\}@mail.utoronto.ca, chiheb.trabelsi@utoronto.ca, } \\
{\tt\small zhibozhang@cs.toronto.edu, j.jang@lgresearch.ai, ssanner@mie.utoronto.ca} \\
}

\maketitle

%%%%%%%%% ABSTRACT
\begin{abstract}
   Weakly supervised semantic segmentation (WSSS) with only image-level supervision is a challenging task. Most existing methods exploit Class Activation Maps (CAM) to generate pixel-level pseudo labels for supervised training. However, due to the local receptive field of Convolution Neural Networks (CNN), CAM applied to CNNs often suffers from partial activation --- highlighting the most discriminative part instead of the entire object area. In order to capture both local features and global representations, the Conformer has been proposed to combine a visual transformer branch with a CNN branch. In this paper, we propose TransCAM, a Conformer-based solution to WSSS that explicitly leverages the attention weights from the transformer branch of the Conformer to refine the CAM generated from the CNN branch. TransCAM is motivated by our observation that attention weights from shallow transformer blocks are able to capture low-level spatial feature similarities while attention weights from  deep transformer blocks capture high-level semantic context. Despite its simplicity, TransCAM achieves a new state-of-the-art performance of 69.3\% and 69.6\% on the respective PASCAL VOC 2012 validation and test sets, showing the effectiveness of transformer attention-based refinement of CAM for WSSS. Code is available at \href{https://github.com/liruiwen/TransCAM}{https://github.com/liruiwen/TransCAM}.
\end{abstract}

%%%%%%%%% BODY TEXT
\section{Introduction}

Semantic segmentation is a fundamental computer vision task that assigns semantic labels to every pixel in an image. Recent years have seen significant progress in deep-learning-based semantic segmentation methods \cite{chen2017deeplab,long2015fully,badrinarayanan2017segnet}. However, these methods require an enormous amount of training data with pixel-level class labels. Such data needs massive fine-grained annotations and is expensive to obtain; hence much effort has been devoted to weakly supervised semantic segmentation (WSSS) that intends to train a segmentation model with weak supervision such as bounding boxes \cite{dai2015boxsup,khoreva2017simple}, scribbles \cite{lin2016scribblesup}, points \cite{bearman2016s}, or image-level class labels \cite{wang2020self,chang2020weakly,zhang2021complementary,ahn2019weakly}. The image-level class label is the most accessible form of annotation since it already exists in large-scale datasets like ImageNet \cite{russakovsky2015imagenet}. Therefore, we focus on WSSS using only image-level class labels in this work.

\begin{figure}[htbp!]
    \centering
    \bgroup
    \def\arraystretch{0.3}
    \setlength\tabcolsep{0.5pt}
    \begin{tabular}{cccc}
        \includegraphics[width=0.24\linewidth]{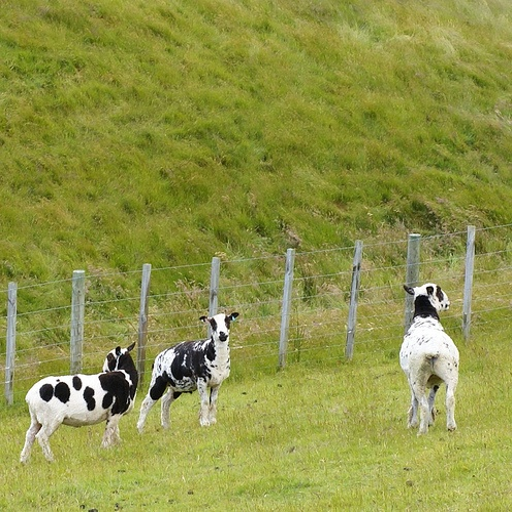} &
        \includegraphics[width=0.24\linewidth]{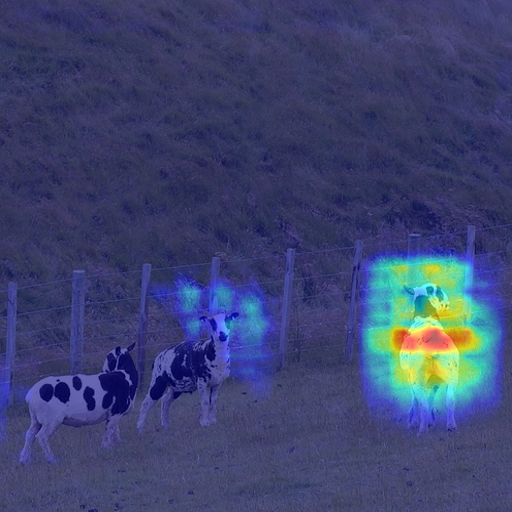} &
        \includegraphics[width=0.24\linewidth]{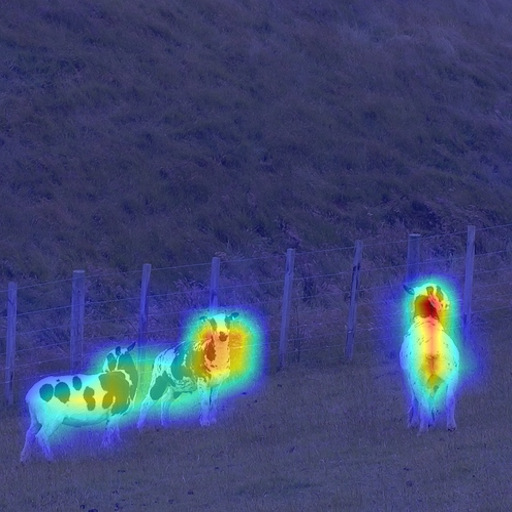} &
        \includegraphics[width=0.24\linewidth]{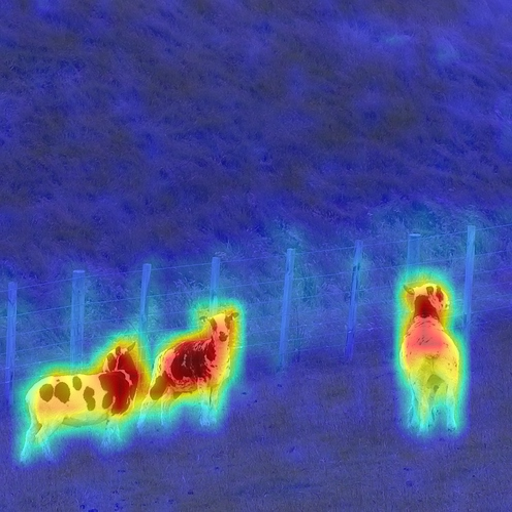} \\
        
        \includegraphics[width=0.24\linewidth]{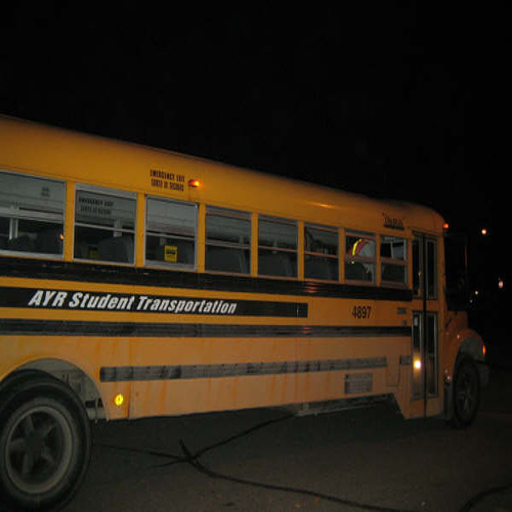} &
        \includegraphics[width=0.24\linewidth]{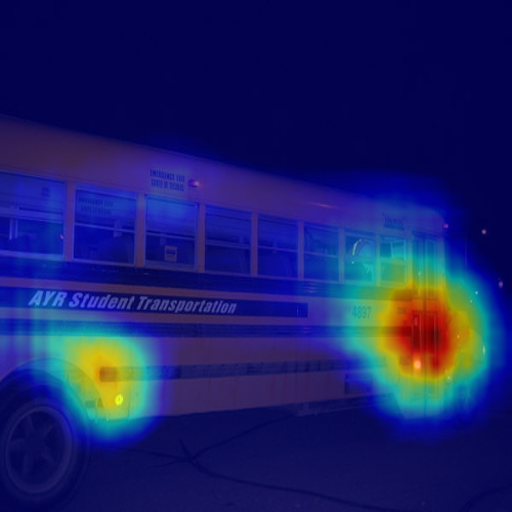} &
        \includegraphics[width=0.24\linewidth]{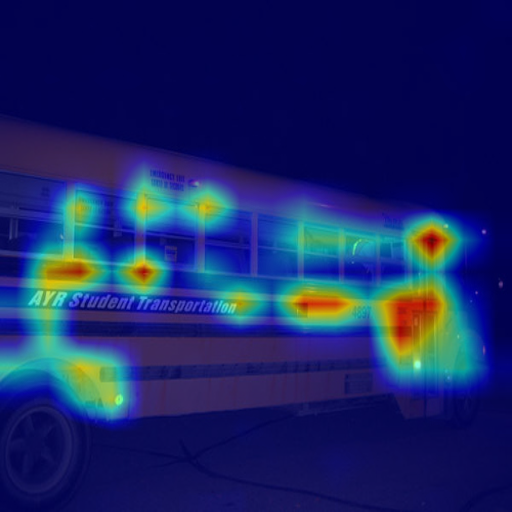} &
        \includegraphics[width=0.24\linewidth]{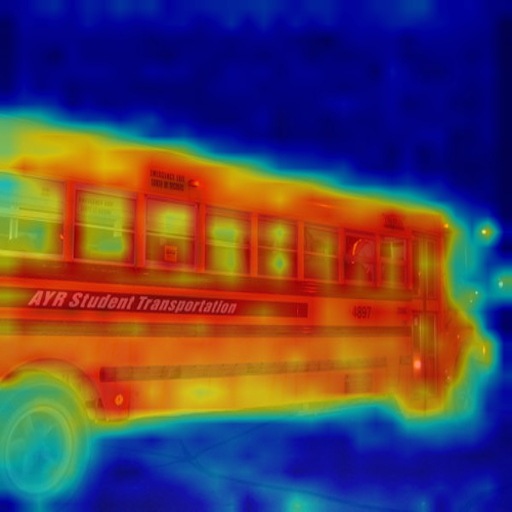} \\
        
        \includegraphics[width=0.24\linewidth]{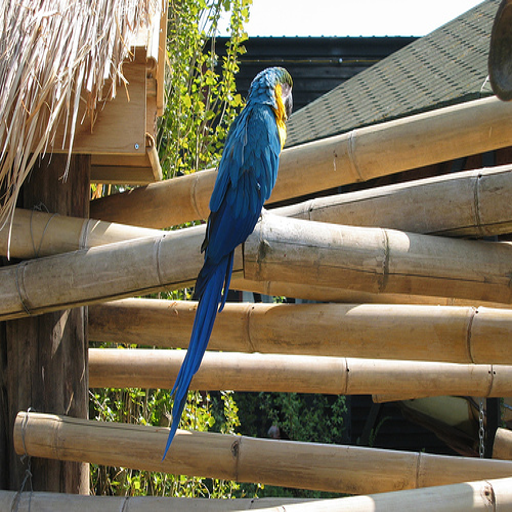} &
        \includegraphics[width=0.24\linewidth]{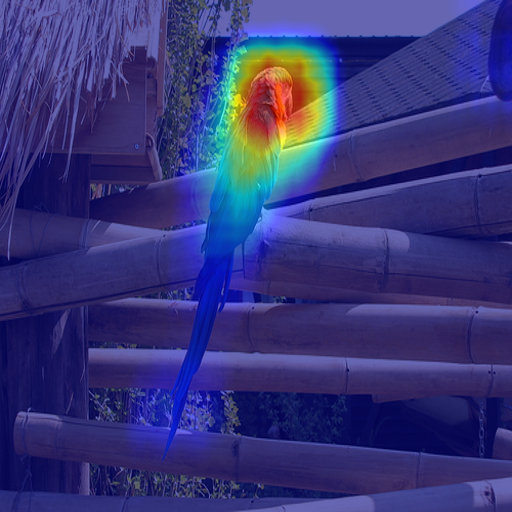} &
        \includegraphics[width=0.24\linewidth]{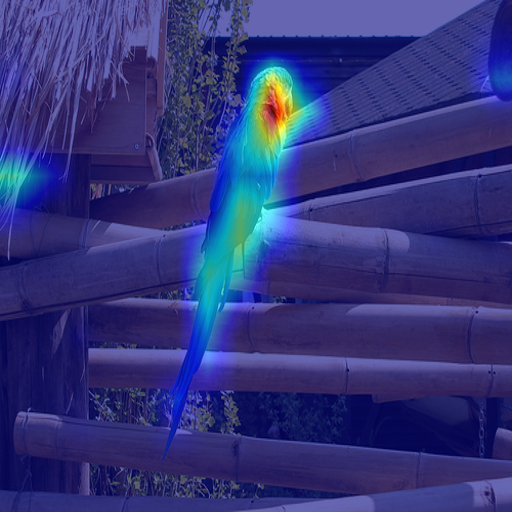} &
        \includegraphics[width=0.24\linewidth]{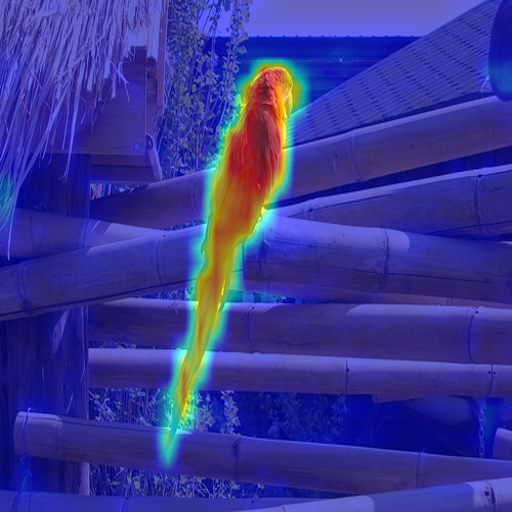} \\
        \small{(a) image} & \small{\makecell[t]{(b) ResNet- \\ CAM}} & \small{\makecell[t]{(c) Conformer- \\ CAM}} & \small{\makecell[t]{(d) TransCAM \\ (ours)}} \\
    \end{tabular}
    \egroup
    \caption{Comparisons of CAMs generated by different networks or methods: (a) the input images, (b) conventional CAMs generated from a ResNet-38 network (c) conventional CAMs generated from the CNN branch of a Conformer-S network (d) CAMs generated by our method.}
    \label{fig:intro}
\end{figure}

To tackle WSSS with only image-level labels, most existing approaches rely on Class Activation Maps \cite{zhou2016learning} (CAM) to generate the initial seeds for localization. Specifically, these methods often train a CNN classification network with image-level labels, from which CAM are generated and refined as pixel-level pseudo labels. Then a fully-supervised semantic segmentation network is trained using the pseudo labels as ground-truth annotations. Clearly, the success of WSSS is critically dependent on the accuracy of the pseudo labels. However, a major obstacle of CAM-derived pseudo labels for WSSS is the partial activation issue --- CAM generated by the classification model tend to highlight the most discriminative part of an object instead of the entire object area \cite{ahn2018learning,wang2020self,chang2020weakly,lee2021anti} as demonstrated in Figure~\ref{fig:intro} (b). Recent work has pointed out the root cause of this issue is an intrinsic characteristic of CNNs: the local receptive field only captures small-range feature dependencies \cite{gao2021ts,chen2021lctr}. Although various methods have been proposed to encourage a larger activation area aligned with the entire object region \cite{li2018tell,wang2020self,wei2017object,ahn2018learning,lee2021anti,lee2019ficklenet}, little work has directly addressed the local receptive field deficiencies of the CNN when applied to WSSS.

Recently, transformers~\cite{vaswani2017attention} have demonstrated tremendous success in many computer vision tasks~\cite{dosovitskiy2020image,touvron2021training}, but they are much less explored for WSSS. The Vision Transformer (ViT) \cite{dosovitskiy2020image} leverages multi-head self-attention and multi-layer perceptrons to capture long-range semantic correlations, which are crucial for localizing the entire object; however, in contrast to the CNN, the ViT often ignores local feature details of objects that are also important for WSSS applications. In order to leverage advantages of both the CNN and ViT architectures, hybrid combinations of the two have been developed \cite{wu2021cvt,peng2021conformer,dai2021coatnet,chen2021mobile,xu2021vitae}. In particular, the Conformer \cite{peng2021conformer} utilizes a CNN branch and a transformer branch to fuse local features and global representations interdependently at multiple scales. 

As a first contribution of our work, we show the Conformer is more effective as a backbone for WSSS when compared to commonly used ResNets \cite{he2016deep,wu2019wider}. As shown in Figure~\ref{fig:intro} (b) \& (c), the Conformer generates CAM with more extensive coverage on the target object and  more accurate pseudo labels from CAM.  This leads to better WSSS performance as shown in Table~\ref{tab:abl_bk}. We believe this improvement derives from the Conformer's ability to couple local features with global representations, which is important to alleviate the partial activation issue of CAM. Nonetheless, CAM generated from Conformer's CNN branch are only influenced by the self-attention weights from the transformer branch in an \textit{implicit} manner (through the Feature Coupling Units (FCU)), as shown in Figure~\ref{fig:method}, leading to poor performance on the WSSS task (Section~\ref{ssec:ablation}).

To address this issue in a second contribution of our work, we propose a simple, efficient, and effective method called \textbf{TransCAM} that explicitly leverages the attention weights from the transformer branch of the Conformer to refine the CAM generated from the CNN branch. TransCAM is motivated by our observation that attention weights from the shallow transformer blocks are able to capture low-level spatial feature similarities while attention weights from deep transformer blocks capture high-level semantic context (Section~\ref{sec: attention_analysis}). Based on this observation, we construct an attention map that embeds both low-level and high-level spatial feature affinities by averaging the multi-head self-attention weights from all transformer blocks. The CAM generated from the CNN branch is refined by this attention map with a simple dot product. As illustrated in Figure~\ref{fig:method}, TransCAM does not rely on any additional components nor complex calculations. It simply leverages the feature affinities embedded in the Conformer backbone and enhances the CAM with a single-step dot product refinement. Thorough experiments demonstrate the effectiveness of TransCAM --- despite its relative simplicity, TransCAM achieves a new state-of-the-art performance of 69.3\% and 69.6\% on the respective PASCAL VOC 2012~\cite{everingham2010pascal} validation and test sets.

We summarize our main contributions:
\begin{itemize}
\item We propose the first transformer-based solution (TransCAM) for WSSS to address the partial activation limitations of CAM due to the inherent deficiencies of the CNN's local receptive field.

\item TransCAM explicitly utilizes the inherent attention weights from the transformer branch of the Conformer to improve the CAM generated from the CNN branch. It is based on our observation that the attention weights capture both low-level and high-level spatial feature affinities. TransCAM does not rely on any additional components nor complex calculations, thus also making it a simple high-performance baseline for further research.

\item TransCAM achieves state-of-the-art performance for WSSS on the PASCAL VOC 2012 dataset with image-level annotations.

\end{itemize}

\section{Related Work}

\subsection{Weakly Supervised Semantic Segmentation (WSSS)}

Most recent solutions to WSSS with image-level annotations use Class Activation Maps (CAM) \cite{zhou2016learning} to generate the pixel-level pseudo labels and then train the segmentation model with the pseudo labels in a fully supervised manner. CAM combines the feature map before the global average pooling layer with the weights of the last fully connected layer to score and visualize the contribution of each pixel to the classification result. However, it tends to focus on the most discriminative part of the target object. 

Existing works that deal with this issue are mainly divided into two categories. The first category makes the network focus on more complete object regions. Adversarial erasing \cite{wei2017object,hou2018self} removes the most discriminative part of an object to encourage larger activation on non-discriminative object regions; Sun \textit{et al.} \cite{sun2020mining} explores cross-image semantic relations, while Chang \textit{et al.} leverages a sub-category objective for object pattern mining; SEAM \cite{wang2020self} exploits consistency regularization for CAM equivalence under affine transformations; AdvCAM \cite{lee2021anti} generates an attribution map from an image that is manipulated against adversarial attack. The second category refines the initial pseudo labels for segmentation via post-processing. SEC \cite{kolesnikov2016seed} refines CAM following three principles: seed, expand, and constrain; AffinityNet \cite{ahn2018learning} uses a network to predict the semantic affinity between neighboring pixels and then uses the prediction result for semantic propagation via random walk; IRNet \cite{ahn2019weakly} further explores the boundary activation map for determining pixel affinity; AuxSegNet \cite{xu2021leveraging} adopts auxiliary tasks and learns cross-task affinity to propagate the CAM.  Despite the collective progress made by these works, little research has directly addressed the local receptive field deficiencies of the CNN when applied to WSSS, which is the key problem that we tackle in this work.

\subsection{Self-attention for WSSS}
The self-attention mechanism \cite{vaswani2017attention} has substantially improved various computer vision tasks including semantic segmentation. Non-local neural networks \cite{wang2018non} apply it to convolutional neural networks to capture long-range feature dependencies. Some recent WSSS works also adopt self-attention modules to capture semantic affinities. CIAN \cite{fan2020cian} leverages self-attention to obtain more extensive and consistent activation regions from the activation maps of two images with the same class objects. SEAM \cite{wang2020self} adopts self-attention to capture pixel affinity that can be used to refine CAM. EDAM \cite{wu2021embedded} applies self-attention to model the intra-image and inter-image homogeneity. However, these methods require attaching extra components to the end of the backbone output, which increases the model complexity and ignores low-level semantic correlations. Our proposed method leverages the inherent multi-head attention weights in the transformer blocks to capture low- and high-level feature affinities.

\subsection{Transformer-based Weakly Supervised Object Localization}
The transformer \cite{vaswani2017attention} architecture is initially proposed to handle sequential data for natural language processing tasks. The Vision Transformer (ViT) \cite{dosovitskiy2020image} splits an input image into sequential patches and feeds them into the original transformer encoder to obtain the visual representation for the image classification task. TS-CAM \cite{gao2021ts} applies the visual transformer to the weakly supervised object localization task by coupling the patch token with the semantic-agnostic attention map. Although the visual transformer captures global representations, it ignores local feature details. To this end, LCTR \cite{chen2021lctr} utilizes cross-patch information to enhance the local perception capability while retaining global information.  To date, none of these methods have been extended to WSSS.

\section{Methodology}
\begin{figure*}[tb]
    \centering
    \includegraphics[width=\textwidth]{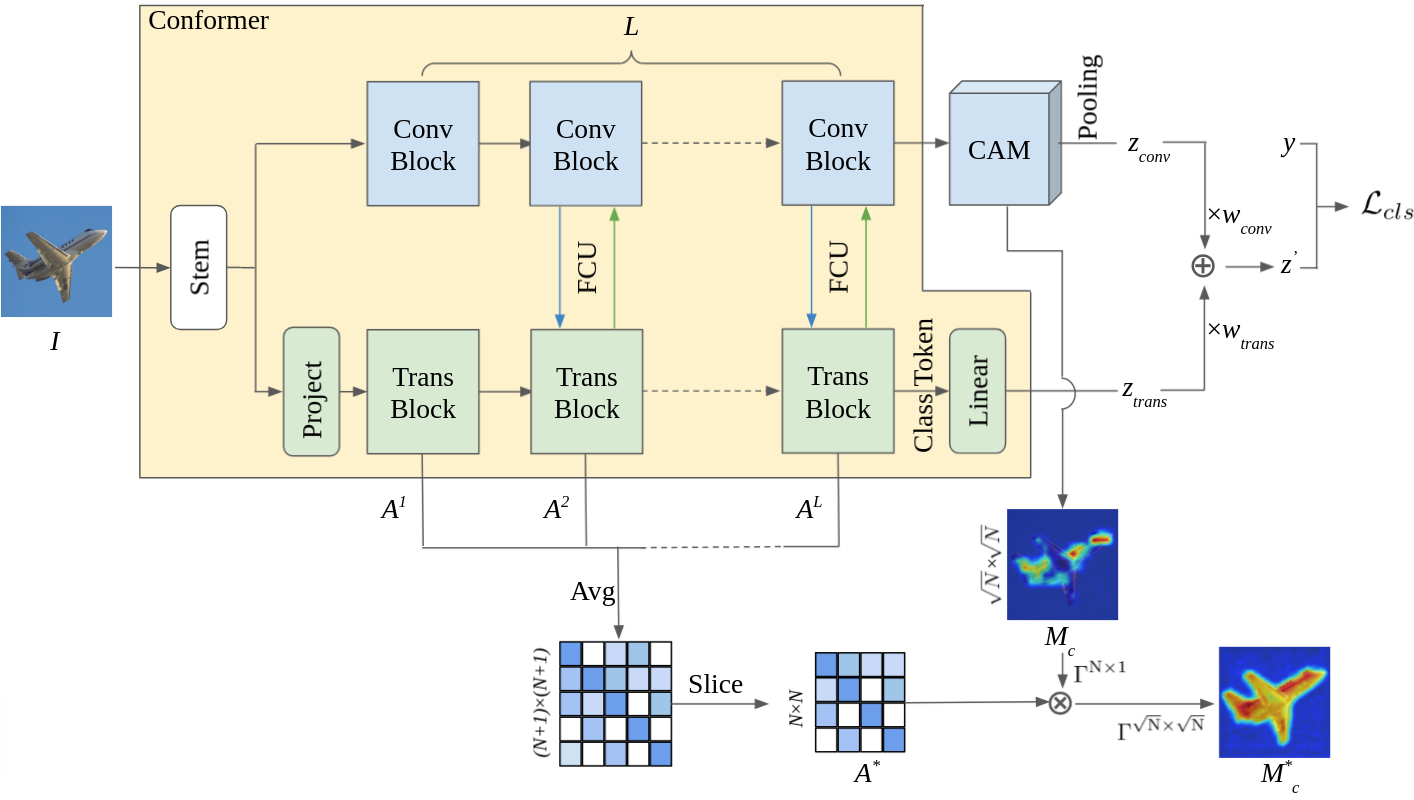}
    \caption{TransCAM leverages the dual-branch Conformer network (yellow part) with $L$ convolution and transformer blocks. It refines the CAM $M_c$ generated from the CNN branch with the Attention Map $A^*$, and the resulting CAM $M_c^*$ has significantly improved object coverage. The Attention Map $A^*$ is constructed by averaging the multi-head self-attention weights from all transformer blocks and excluding all the class token weights (slice). TransCAM does not rely on any additional components nor complex calculations. It simply leverages the low-level and high-level spatial feature affinities embedded in the Conformer backbone to enhance the CAM with a single-step dot product refinement.}
    \label{fig:method}
\end{figure*}

\subsection{Preliminaries}

\subsubsection{The Conformer network} The Conformer  \cite{peng2021conformer} is a dual network structure aiming to couple CNN-based local features with transformer-based global representations for enhanced representation learning, as shown in Figure~\ref{fig:method} (yellow part). Specifically, the Conformer consists of a stem module to extract initial local features, a CNN branch, a transformer branch, 
and Feature Coupling Units (FCUs) to progressively fuse local features in the CNN branch with global representations in the transformer branch. Note that the CNN and transformer branches have the same number of ($L$) repeated convolution and transformer blocks, respectively. 

The first transformer block projects the output from the stem module using a convolution to generate the patch embeddings. Then, the patch embeddings are prepended with a class token, used as the input for the multi-head self-attention (MHSA) module. As for later transformer blocks, it takes the output of the FCU and adds to the token embeddings from the previous transformer block; the result concatenated with the class token is used as the input for the MHSA module. We denote the input and the output for the MHSA module in the $l$-th block by $X^l \in \mathbb{R}^{(1+N)\times D}$ and $\tilde{X}^l \in \mathbb{R}^{(1+N)\times D}$, where $1$, $N$, $D$ represents the class token, the spatial dimension of token embeddings, and transformer embedding dimensions, respectively. 

The CNN branch adopts a feature pyramid structure, where the resolution of feature maps decreases with network depth while the channel number increases. Each convolution block in the CNN branch contains multiple bottlenecks, which comprises a down-projection convolution, a spatial convolution, and an up-projection convolution, following the definitions in ResNet~\cite{he2016deep}.

\subsubsection{Class Activation Map} A Class Activation Map (CAM)~\cite{zhou2016learning} for a particular class indicates the discriminative image regions used by the CNN to identify that class, which is often used to generate pseudo labels in WSSS. It can be applied to a trained classification network that contains a feature extracting network, a Global Average Pooling (GAP) layer, and a linear classification layer. The activation map is generated by multiplying the feature maps from the last convolution layer with the target-specific weights of the classification layer. The generation of the activation map that is related to the $c$-th class can be formulated as:
\begin{equation}
    M_c = \theta_c^{\intercal}f
\end{equation}
where $M_c\in \mathbb{R}^{f_h\times f_w}$ is the class-specific activation map, $\theta_c \in \mathbb{R}^{f_c\times 1}$ are the weights of the classification layer that correspond to the $c$-th object, and $f \in \mathbb{R}^{f_c \times f_h \times f_w}$ is the feature map from the last convolution layer ($f_c, f_h, f_w$ are the number of channels, the height, and the width of the feature map respectively). For WSSS, one can also implement CAM using fully convolutional networks without the linear classification layer \cite{hwang2016self}. 

\subsection{CAM Generation from Conformer}
We adopt the Conformer as the backbone network for the initial pseudo label generation. For the original Conformer network, the last bottleneck of the final convolution has a spatial convolution with stride 2 for down-scaling. We modify it with stride 1 to get higher resolution feature maps for better segmentation performance and match the size of the Attention Map described in Section \ref{sec:coupling}. After the modification, the feature maps $f$ from the CNN branch are of the size $f_c\times\sqrt{N}\times\sqrt{N}$ where $f_c$ is the number of channels, and $N$ denotes the spatial dimension of token embeddings. 

CAM is an efficient method to localize objects given image class labels. We use CAM to generate the activation map $M \in \mathbb{R}^{C\times\sqrt{N}\times\sqrt{N}}$ from the CNN branch of the Conformer where C and N are the respective number of classes (including the background class) and spatial dimension of token embeddings. As shown in Figure~\ref{fig:intro} (b) \& (c), the CAM from the Conformer highlights more extensive object regions compared to the CAM from conventional CNNs \cite{simonyan2014very,he2016deep} since the Conformer employs the MHSA mechanism with the global receptive field to capture long-range feature dependencies. However, it is still hard to achieve ideal performance for the WSSS task if we directly use CAM from the Conformer for pseudo label generation since the CAM is only influenced by the self-attention weights from the transformer branch in an \textit{implicit} manner (through FCU) and does not fully exploit the feature affinity information from the attention weights for localization.

\subsection{Attention Map Generation}
We assume that the weights of the multi-head self-attention module in the $l$-th transformer block are $A^l \in \mathbb{R}^{S\times(1+N)\times(1+N)}$, where $S$ is the number of heads. The attention weights $A^l$ are computed as:
\begin{equation}
    A^l = \mathrm{softmax}(\frac{Q^l {K^l}^{\intercal}}{\sqrt{D/S}})
\end{equation}

where $Q^l$ and $K^l$ are the key and query representations projected from $X^l$ in the $l$-th transformation block. To aggregate the semantic context captured by all heads, we take the average of $A^l$ across all heads to get $\bar{A^l} \in \mathbb{R}^{(1+N)\times(1+N)}$. $\bar{A^l}_{i,j} \in \mathbb{R}$ reveals how much the $j$-th input token embedding contributes to the $i$-th output token embedding of the MHSA module in the $l$-th transformer block, and $\bar{A^l}$ (excluding the weights for the class token) captures spatial feature affinities in $l$-th transformation block in general. 

\begin{figure}[htb!]
\centering
    \bgroup
    \def\arraystretch{0.3}
    \setlength\tabcolsep{0.5pt}
    \begin{tabular}{ccccc}
        \includegraphics[width=0.24\linewidth]{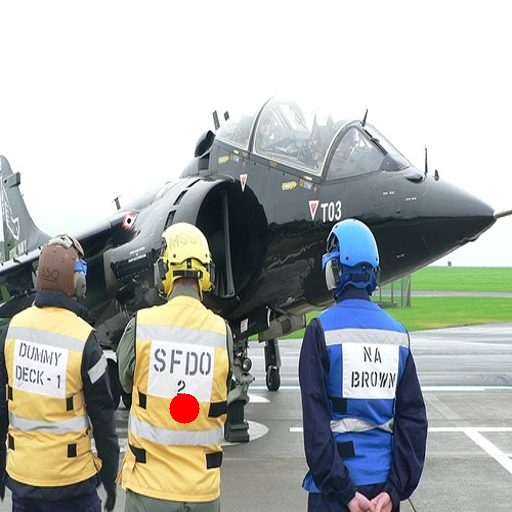} &
        \includegraphics[width=0.24\linewidth]{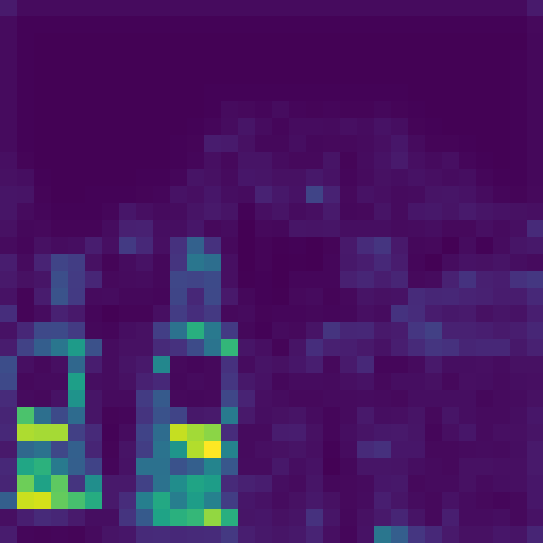} &
        \includegraphics[width=0.24\linewidth]{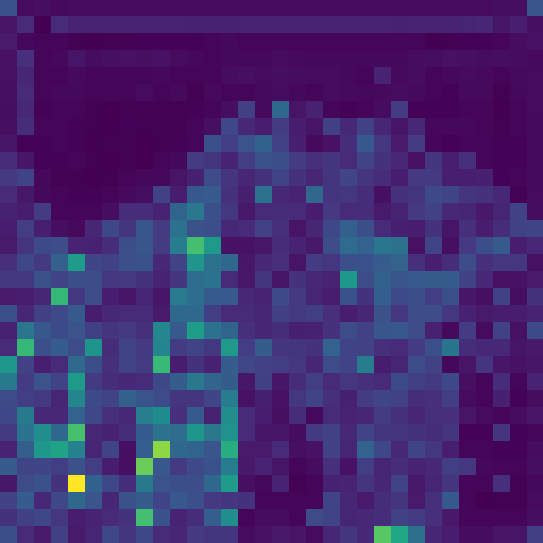} &
        \includegraphics[width=0.24\linewidth]{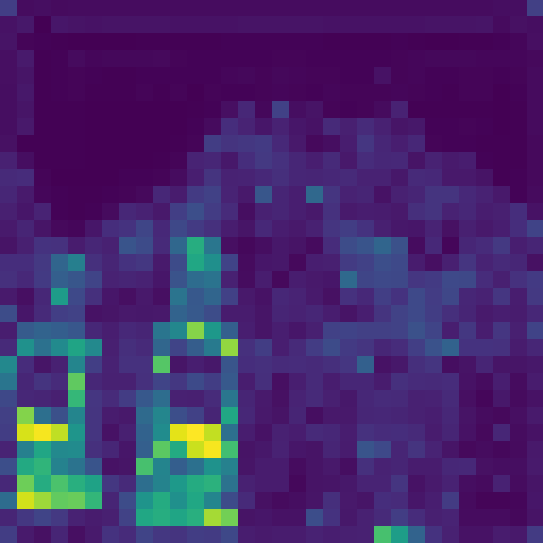} \\
        \includegraphics[width=0.24\linewidth]{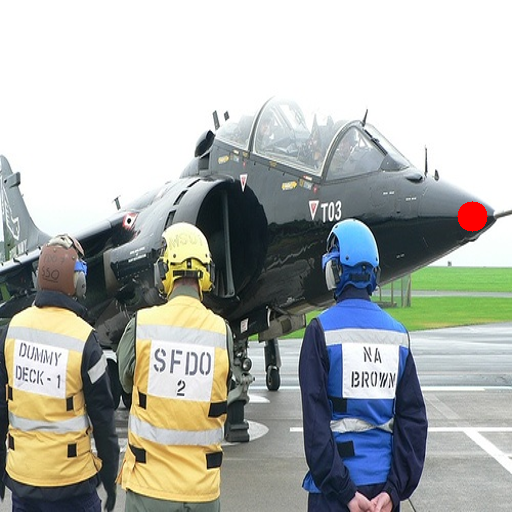} &
        \includegraphics[width=0.24\linewidth]{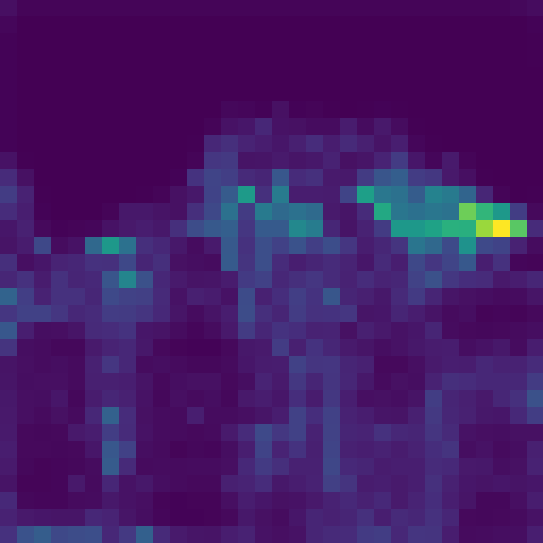} &
        \includegraphics[width=0.24\linewidth]{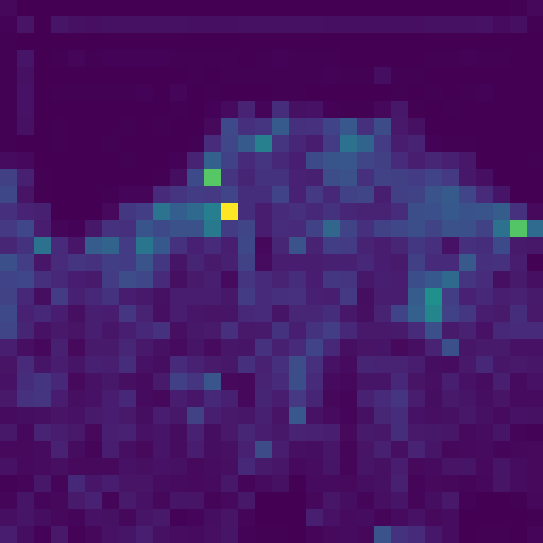} &
        \includegraphics[width=0.24\linewidth]{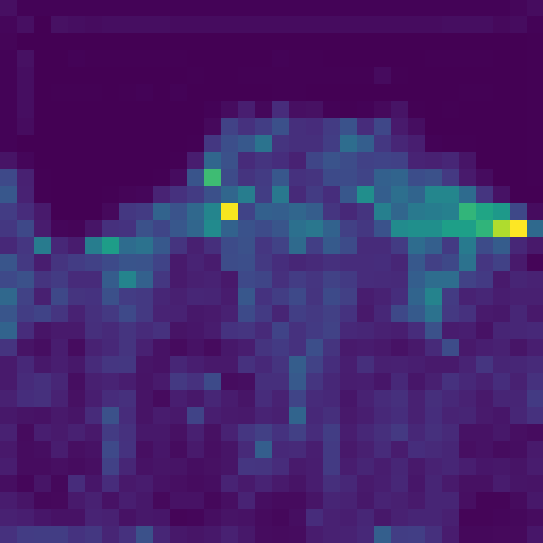} \\
        \includegraphics[width=0.24\linewidth]{figures/fig3/2007_001825_1.png} &
        \includegraphics[width=0.24\linewidth]{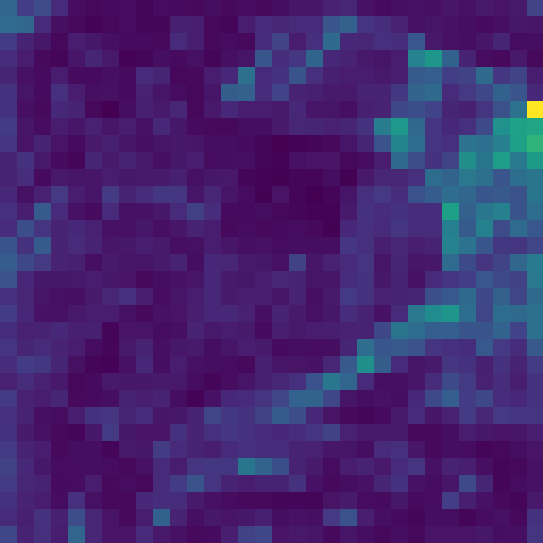} &
        \includegraphics[width=0.24\linewidth]{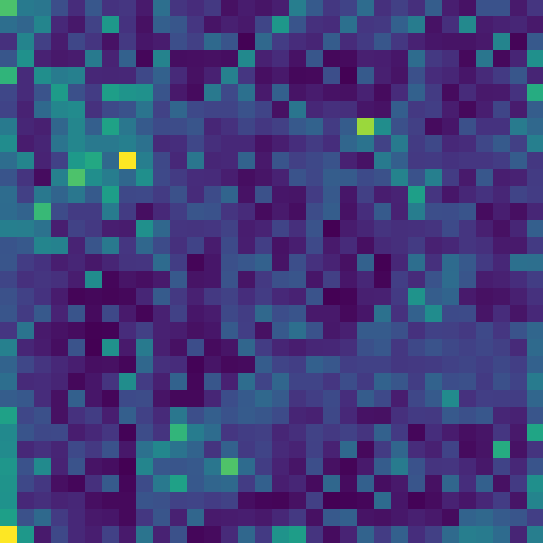} &
        \includegraphics[width=0.24\linewidth]{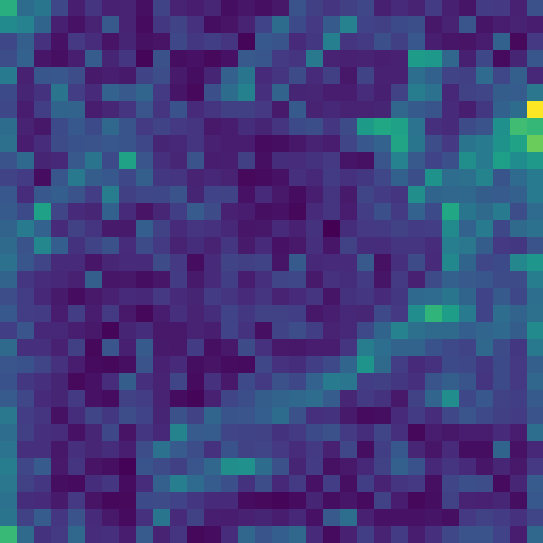} \\
        \includegraphics[width=0.24\linewidth]{figures/fig3/2007_001825_2.png} &
        \includegraphics[width=0.24\linewidth]{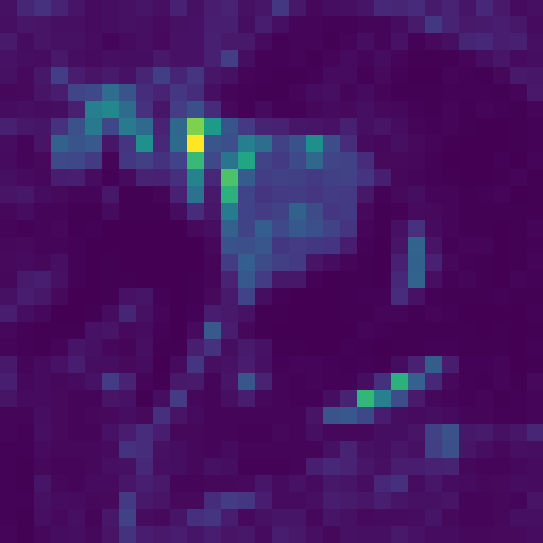} &
        \includegraphics[width=0.24\linewidth]{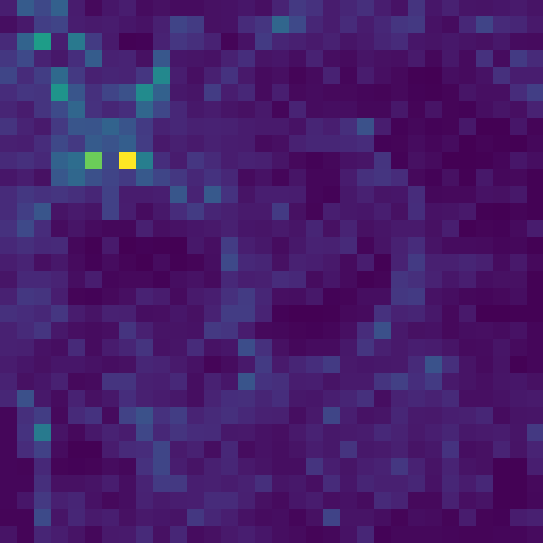} &
        \includegraphics[width=0.24\linewidth]{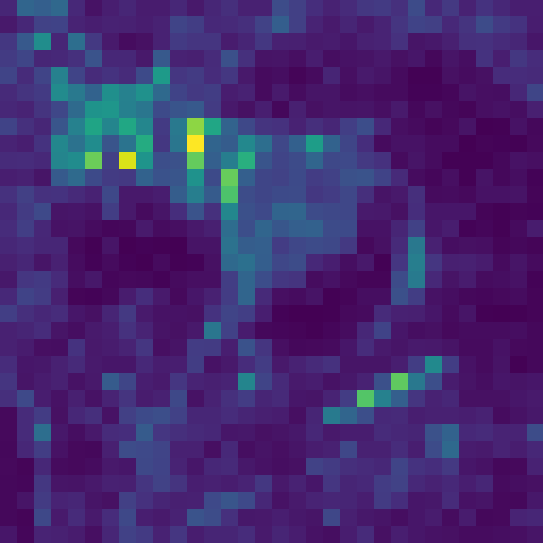} \\
        \includegraphics[width=0.24\linewidth]{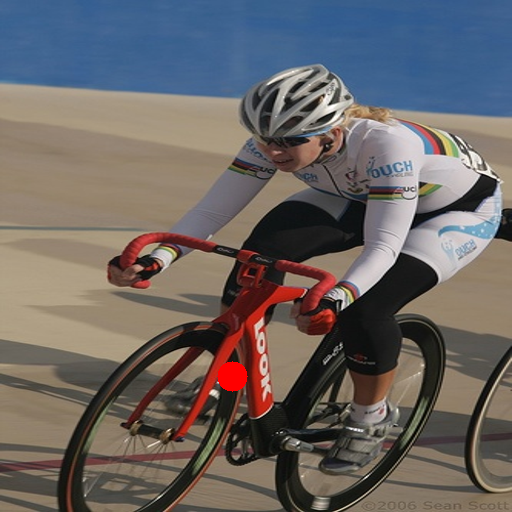} &
        \includegraphics[width=0.24\linewidth]{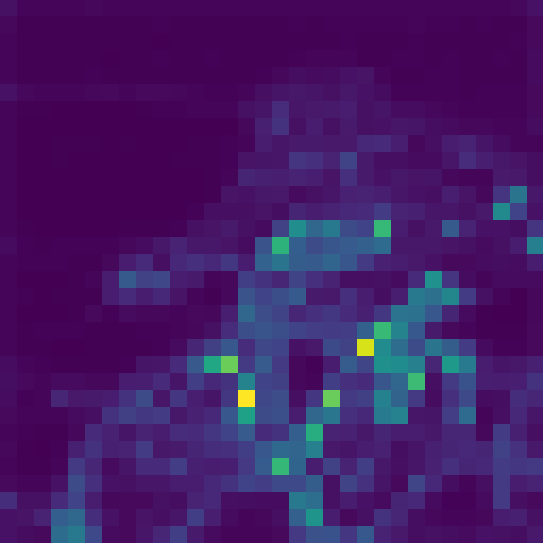} &
        \includegraphics[width=0.24\linewidth]{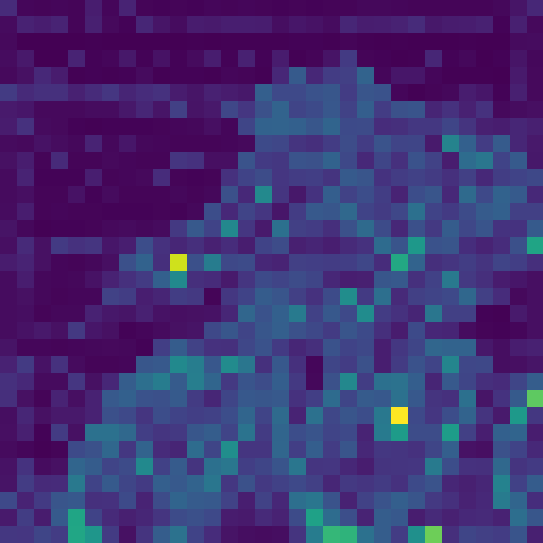} &
        \includegraphics[width=0.24\linewidth]{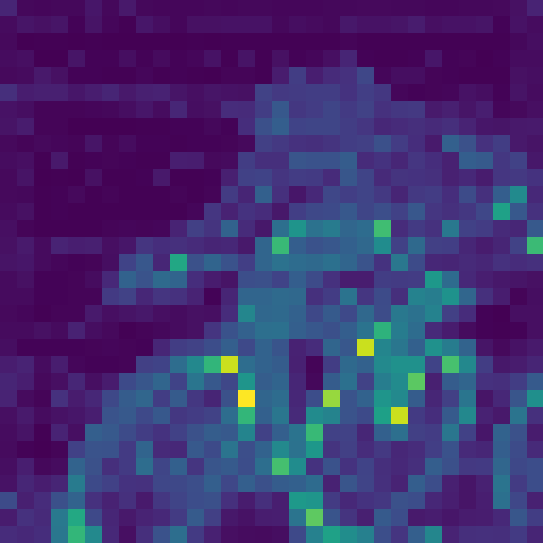} \\
        \includegraphics[width=0.24\linewidth]{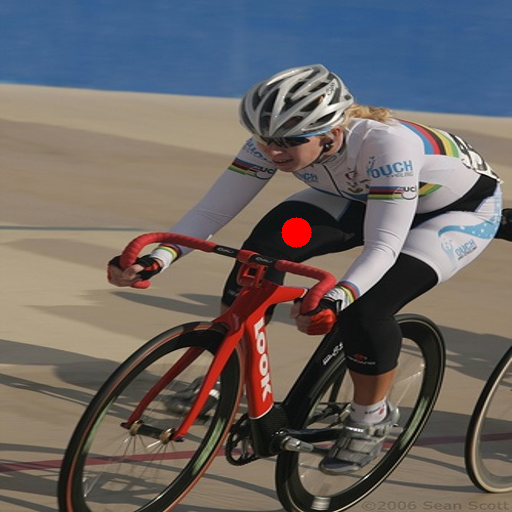} &
        \includegraphics[width=0.24\linewidth]{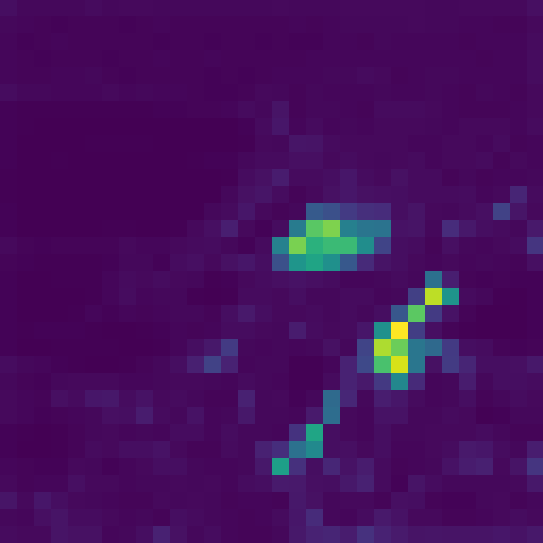} &
        \includegraphics[width=0.24\linewidth]{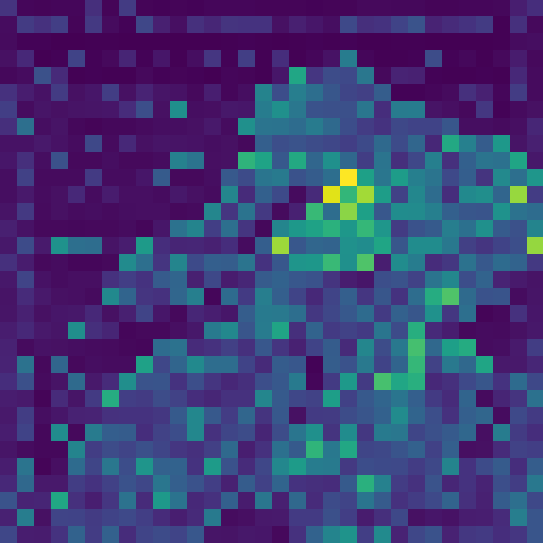} &
        \includegraphics[width=0.24\linewidth]{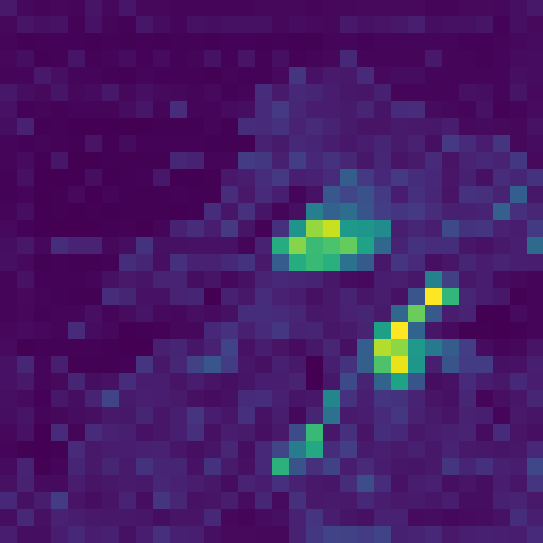} \\
        \\
        (a) images & (b) AS & (d) AD & (e) AA\\
    \end{tabular}
    \caption{Visualization of input images with reference points (red dots) and the Attention Maps of the reference points. We show the Attention Maps generated by different weighting methods: averaging attention weights in shallow blocks (AS), deep blocks (AD), and all blocks (AA).}
    \label{fig:abl_aw}
    \egroup
\end{figure}

In order to acquire a single attention map, we can compute the average attention weights $\bar{A} \in \mathbb{R}^{(1+N)\times(1+N)}$ over different transformer blocks:
\begin{equation}
    \bar{A} = \frac{1}{L}\sum_{l}{\bar{A^l}}
\end{equation}
To better understand what information the attention weights from different blocks capture, we compare the average attention weights in shallow blocks (first half), denoted by \textit{AS}; deep blocks (second half), denoted by \textit{AD} and all blocks, denoted by \textit{AA} in Figure~\ref{fig:abl_aw}. It can be observed that \textit{AS} tends to assign high weights to the pixels that are similar in texture to the reference point, which correlates with the fact that shallow transformer blocks leverage the local feature captured by the early convolution blocks. \textit{AD} reveals the ability of the Attention Map to capture high-level semantics, \textit{e.g.}, it manages to capture three persons instead of two for \textit{AS} in the first image; it highlights the eyes and ears of the cat in the fourth image. However, it looks noisy because token embeddings in deep layers reflect complex semantic context. \textit{AA} is able to leverage both low-level local feature similarities from shallow blocks and high-level semantic context from deep blocks. It can also smooth the noise in the attention weights from deep blocks. Based on this observation,  we believe that it is crucial to combine both low-level and high-level feature affinities for CAM refinement, and therefore, we compute the average attention weights $\bar{A}$ over all transformer blocks.

We ignore the attention weights in $\bar{A}$ that are related to the class token since they only imply the contribution of each token embedding to the classification result but do not reflect the spatial feature affinities. The final Attention Map $A^* \in \mathbb{R}^{N\times N}$ can be computed as:
\begin{equation}
    A^* = (\bar{A}_{i,j})_{2\leq i \leq N+1, 2\leq j \leq N+1}
\end{equation}

\subsection{Attention-based CAM refinement}
\label{sec:coupling}
The attention weights contain semantic information that could be leveraged to generate CAM with extensive object coverage. Multiple methods could be used to fuse the semantic information with CAM as shown in Section \ref{ssec:ablation}. We propose to leverage the spatial feature affinities captured by $A^*$ to refine the CAM $M_c$ via matrix multiplication:
\begin{equation}
    M_c^* = \mathrm{\Gamma^{\sqrt{N}\times\sqrt{N}}}(A^* \cdot \mathrm{\Gamma^{N\times1}}(M_c))
\end{equation}
where $M_c^* \in \mathbb{R}^{\sqrt{N}\times\sqrt{N}}$ is the refined CAM, $\mathrm{\Gamma^{N\times1}}$ is the reshape operator that flattens $M_c$ to match the size of the patch embeddings, and $\mathrm{\Gamma^{\sqrt{N}\times\sqrt{N}}}$ is the operator that reshapes to the dimension of $M_c$. $M_c^*$ explicitly exploits the context information from the transformer attention weights and can be used to generate an initial pseudo label with extensive object coverage for the WSSS task.

\subsection{Training and Pseudo Label Generation}
We apply the GAP layer to the activation map $M$ from the CNN branch to obtain the prediction vector $z_{conv} \in \mathbb{R}^C$. Following the original Conformer, we attach a linear layer to the class token at the end of the transformer branch and obtain the prediction vector $z_{trans} \in \mathbb{R}^C$. We use the weighted sum of $z_{conv}$ and $z_{trans}$ as the prediction logits: $z' = w_{conv} z_{conv} + w_{trans} z_{trans}$, where $z' \in \mathbb{R}^C$, $w_{conv} \in \mathbb{R}$, $w_{trans} \in \mathbb{R}$ denotes the final logits, the weight for convolution logits, the weight for transformer logits, respectively. We employ multi-label soft margin loss for network training. The classification loss defined for $C - 1$ foreground image labels is formulated as:
\begin{equation}
    \mathcal{L}_{cls} = - \frac{1}{C-1}\sum_{c=1}^{C-1}[y_c\mathrm{log}(\frac{1}{1+e^{-z'_c}}) + (1 - y_c)\mathrm{log}(\frac{e^{-z'_c}}{1+e^{-z'_c}})]
\end{equation}
where $z'_c \in \mathbb{R}$ is the logit of the $c$-th object, and $y_c \in \{0,1\}$ is the ground truth class annotation of the $c$-th object.
After we have the trained classification model, to generate the pseudo labels, we set the background score as:
\begin{equation}
    M^{bkg}_{h, w} = \tau
\end{equation}
where $\tau \in [0, 1]$ denotes the hard threshold parameter at the position ($h$, $w$), $1 \leq h \leq H$, $1 \leq w \leq W$, $H$ and $W$ are the height and the width of the image. During inference, we apply elementwise min-max normalization to all refined foreground CAMs ${M^*_c}_{1\leq c \leq C-1}$ and concatenate them together. We up-sample the normalized result to the size of the input image, then we append $M^{bkg}$ with the result denoted as $\hat{M^*} \in C\times H \times W$. The pixel-level pseudo label is calculated as:
\begin{equation}
    Y_{h, w} = \mathrm{argmax}_c \hat{M^*}_{h,w}
\end{equation}
where $Y_{h, w} \in \{1,...,C\}$ denotes the class label of the pixel at position $h, w$.

\section{Experiments}
\subsection{Dataset and Evaluation Metric}
We evaluate our method on the PASCAL VOC 2012 segmentation benchmark \cite{everingham2010pascal}. There are 21 object categories in the dataset, including 20 foreground objects and the background. The official dataset is split into 1464 images for training, 1449 images for validation, and 1456 images for testing. Following the conventional experimental protocol used in the previous works \cite{wang2020self,ahn2018learning,lee2019ficklenet,jo2021puzzle,zhang2021complementary}, we build an augmented train set with 10582 images by taking additional annotations from the Semantic Boundary Dataset \cite{hariharan2011semantic}. We leverage only image-level class labels during training. To measure the performance of our method, we use the mean Intersection-over-Union (mIoU) as the evaluation metric. 

\subsection{Implementation Details}
We adopt Conformer-S \cite{peng2021conformer} with pre-trained weights on ImageNet \cite{russakovsky2015imagenet} as the backbone network. The Conformer-S network has 37.7M parameters, which is less than ResNet-101 (44.5M). The dual-branch importance weights $(w_{conv}, w_{trans})$ are set to (0.5, 0.5), and the performance of other weights are reported in Section \ref{ssec:ablation}. Our model is trained on a single Tesla V100 GPU with 32GB RAM. We use standard data augmentations during training: random scaling, random horizontal flipping, color jittering, and random cropping. We randomly re-scale the training image in the range of [320, 640] and crop them into the size of 512$\times$512 as the network input. We train our model for 20 epochs with the AdamW \cite{loshchilov2017decoupled} optimizer and batch size 8. The learning rate is set to 0.00005, and the weight decay is set to 0.0005. During inference, aggregating prediction maps from multi-scaled input of a test image is commonly used to boost the final performance. We resize a test image into 256$\times$256, 512$\times$512, and 768$\times$768 during inference.

\subsection{Ablation Studies}
\label{ssec:ablation}
We conduct ablation studies with results validated on PASCAL VOC 2012 \textit{train} set to verify the effectiveness of our proposed approach. In all experiments, the background score $\tau$ is set to the value that gives the best mIoU for the generated pseudo labels on the \textit{train} set in order to ensure a fair comparison.

\begin{table}[htb]
    \centering
    \begin{tabular}{p{0.3\textwidth}l}
        \hline
        method &  mIoU (\%)\\
        \hline
        ResNet-38-CAM\textsuperscript{\textdagger} & 47.16\\
        ResNet-38-CAM \cite{wang2020self} & 47.43\\
        ResNet-50-CAM \cite{ahn2019weakly} & 48.30\\
        Conformer-S-CAM & 51.70\\
        \hline
    \end{tabular}
    \caption{Comparison of pseudo labels produced by CAM with different backbones.
    \textsuperscript{\textdagger}~denotes our implementation.}
    \label{tab:abl_bk}
\end{table}

\subsubsection{CAM with different backbones}
In previous works, CNN models such as ResNet-38 \cite{wu2019wider} and ResNet-50 \cite{he2016deep} are widely used to generate the initial pseudo labels. Table \ref{tab:abl_bk} compares the pseudo labels generated by CAM with ResNet-38, ResNet-50, and Conformer-S \cite{peng2021conformer} backbone. CAM with Conformer-S outperforms CAM with ResNet-50 by 3.4\%, verifying that Conformer generates CAM with more extensive object coverage. Thanks to the interactive fusion of CNN features and transformer embeddings, the Conformer captures local features and global representations; thus it serves as an  effective backbone for the WSSS task.

\subsubsection{Coupling CAM with attention weights} 
We denote our baseline: the CAM from the Conformer-S CNN branch without coupling Attention Map as \textit{Conformer-S-CAM}. We explore different ways to leverage the context information from the attention weights for CAM refinement. One approach is to employ the attention weights $A^{cls}$ for the class token. $A^{cls}$ highlights the crucial pixels for image classification without focusing on local areas. Although $A^{cls}$ is not distinguishable for object classes, we can fuse it with CAM by Hadamard product to make it aware of object categories. We denote this method as \textit{ClsAttn}. Another approach to couple attention weights with CAM is aggregating the patch attention maps based on the patch activation values obtained from CAM, denoted by \textit{AttnAgg}. The detailed calculations and the qualitative results of \textit{ClsAttn} and \textit{AttnAgg} are given in the Appendix. 
Table \ref{tab:abl_bs}
shows the performance comparison of our proposed method with \textit{Conformer-S-CAM}, \textit{ClsAttn}, and \textit{AttnAgg}. Both \textit{ClsAttn} and \textit{AttnAgg} do not have noticeable improvement from the baseline, while our proposed approach increases the mIoU by 13.15\%. It verifies the effectiveness of our method that leverages spatial feature affinities for CAM refinement.

\begin{table}[htbp]
     \centering
     \begin{tabular}{p{0.6\linewidth}l}
         \hline
         method &  mIoU (\%)\\
         \hline
         Conformer-S-CAM (baseline) & 51.70\\
         ClsAttn & 51.15\\
         AttnAgg & 51.87\\
         TransCAM (ours) & 64.85\\
         \hline
     \end{tabular}
     \caption{Comparison with our baseline and different attention-based refinement approaches.}
     \label{tab:abl_bs}
\end{table}

\subsubsection{Leveraging attention weights in all blocks}
\label{sec: attention_analysis}
There are 12 transformer blocks in the Conformer-S network. We compare different methods for generating the final Attention Map $A^*$: averaging attention weights in shallow blocks (Block 1-6), denoted by \textit{AS}; deep blocks (Block 6-12), denoted by \textit{AD}; all blocks (Block 1-12), denoted by \textit{AA}. Table \ref{tab:abl_aw} shows the difference in mIoU of applying these attention weighting methods. \textit{AA} achieves better performance than \textit{AS} and \textit{AD} in terms of mIoU, verifying the importance of combining low-level feature affinities and high-level semantic correlations for CAM refinement.

\begin{table}[htbp]
    \centering
    \begin{tabular}{p{0.4\linewidth}l}
        \hline
        method &  mIoU (\%)\\
        \hline
        AS & 60.39\\
        AD & 63.89\\
        AA & 64.85\\
        \hline
    \end{tabular}
    \caption{Comparison of various attention weighting methods that average attention weights from different transformer blocks.}
    \label{tab:abl_aw}
\end{table}

\subsubsection{Importance weights for dual branches}
\begin{figure}[htbp!]
\centering
\includegraphics[width=\linewidth]{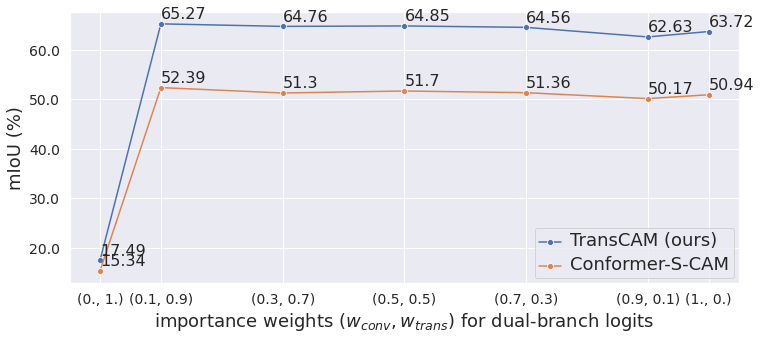}
\caption{Impact of importance weights for dual-branch logits on the pseudo label in terms of mIoU. When $w_{conv}>0.1$, the importance weights have little influence on the pseudo label performance.}
\label{fig:lgs_w}
\end{figure}

We experiment with different combinations of importance weights for logits $(w_{conv}, w_{trans})$, which control the relative importance of the CNN branch and the transformer branch for image classification. As shown in Figure \ref{fig:lgs_w}, logit weights have little impact on the pseudo label quality when $w_{conv}>0.1$ for both our method and our baseline (Conformer-S-CAM). The performance is expected to be low when $w_{conv}$ is 0 since the essential weights for CAM generation are not trained. In other experiments, we report the performance with $(w_{conv}, w_{trans})$ assigned to $(0.5, 0.5)$.

\begin{figure*}[tb!]
    \centering
    \bgroup
    \def\arraystretch{0.2}
    \setlength\tabcolsep{0.5pt}
    \begin{tabular}{m{0.03\linewidth}m{0.95\linewidth}}
        (a) & \includegraphics[width=\linewidth]{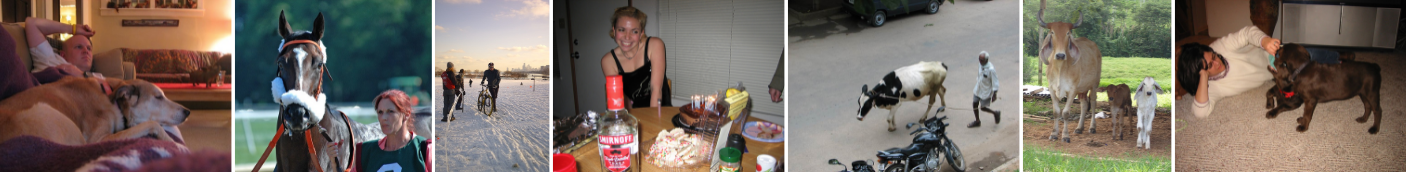}\\
        (b) & \includegraphics[width=\linewidth]{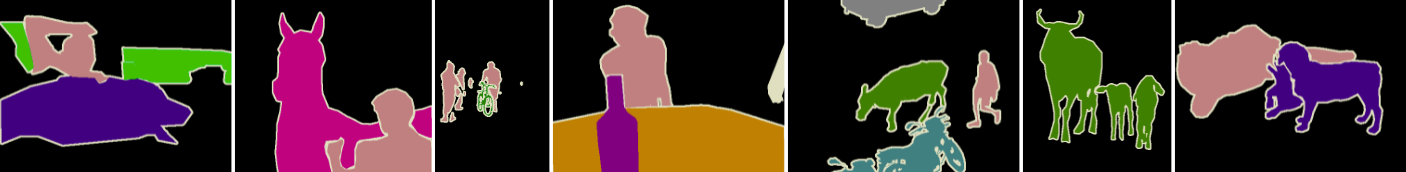}\\
        (c) & \includegraphics[width=\linewidth]{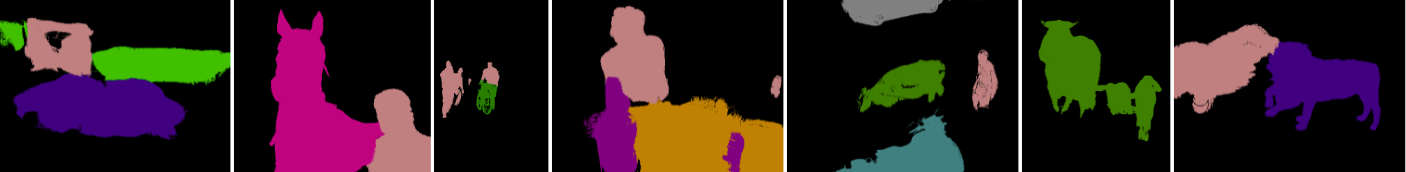}\\
    \end{tabular}
    \caption{Qualitative semantic segmentation results on PASCAL VOC 2012 \textit{val} set. (a) Original images; (b) Ground truth labels; (c) Prediction of the segmentation model trained on  TransCAM generated pseudo labels.}
    \label{fig:qualitative}
    \egroup
\end{figure*} 

\begin{table*}[ht!]
\resizebox{\textwidth}{!}{
\setlength\tabcolsep{1.5pt}
\begin{tabular}{l|ccccccccccccccccccccc|c}
\hline
method & bkg & aero & bike & bird & boat & bottle & bus & car & cat & chair & cow & table & dog & horse & mbk & person & plant & sheep & sofa & train & tv & mIoU\\
\hline
SEC \cite{kolesnikov2016seed} & 82.4 & 62.9 & 26.4 & 61.6 & 27.6 & 38.1 & 66.6 & 62.7 & 75.2 & 22.1 & 53.5 & 28.3 & 65.8 & 57.8 & 62.3 & 52.5 & 32.5 & 62.6 & 32.1 & 45.4 & 45.3 & 50.7\\
AdvErasing \cite{wei2017object} & 83.4 & 71.1 & 30.5 & 72.9 & 41.6 & 55.9 & 63.1 & 60.2 & 74.0 & 18.0 & 66.5 & 32.4 & 71.7 & 56.3 & 64.8 & 52.4 & 37.4 & 69.1 & 31.4 & 58.9 & 43.9 & 55.0\\
PSA \cite{ahn2018learning} & 88.2 & 68.2 & 30.6 & 81.1 & 49.6 & 61.0 & 77.8 & 66.1 & 75.1 & 29.0 & 66.0 & 40.2 & 80.4 & 62.0 & 70.4 & 73.7 & 42.5 & 70.7 & 42.6 & 68.1 & 51.6 & 61.7\\
SEAM \cite{wang2020self} & 88.8 & 68.5 & 33.3 & \textbf{85.7} & 40.4 & 67.3 & 78.9 & 76.3 & 81.9 & 29.1 & 75.5 & 48.1 & 79.9 & 73.8 & 71.4 & 75.2 & \textbf{48.9} & 79.8 & 40.9 & 58.2 & 53.0 & 64.5\\
FickleNet \cite{lee2019ficklenet} & 89.5 & 76.6 & 32.6 & 74.6 & 51.5 & 71.1 & 83.4 & 74.4 & 83.6 & 24.1 & 73.4 & 47.4 & 78.2 & 74.0 & 68.8 & 73.2 & 47.8 & 79.9 & 37.0 & 57.3 & \textbf{64.6} & 64.9\\
Chang \textit{et al.} \cite{chang2020weakly}& 88.8 & 51.6 & 30.3 & 82.9 & 53.0  & \textbf{75.8} & \textbf{88.6} & 74.8 & \textbf{86.6} & \textbf{32.4} & \textbf{79.9} & \textbf{53.8} & 82.3 & \textbf{78.5} & 70.4 & 71.2 & 40.2 & 78.3 & 42.9 & 66.8 & 58.8 & 66.1\\
\hline
\textbf{Ours} & \textbf{91.3} & \textbf{81.9} & \textbf{35.4} & 84.7 & \textbf{67.6} & 67.9 & 87.5 & \textbf{80.5} & 86.5 & 31.4 & 73.9 & 52.5 & \textbf{84.0} & 74.9 & \textbf{74.6} & \textbf{79.0} & 44.7 & \textbf{84.1} & \textbf{47.0} & \textbf{78.4} & 46.6 & \textbf{69.3}\\
\hline
\end{tabular}
}
\caption{Categorical semantic segmentation performance on PASCAL VOC 2012 \textit{val}.}
\label{tab:categorical}
\end{table*}

\begin{table}[tbp!]
    \centering
    \begin{tabular}{p{0.5\linewidth}cc}
        \hline
        Method & w/o p.p. & w/ p.p.\\
        \hline
        $\mathrm{Mixup}$-$\mathrm{CAM\;_{BMVC20}}$ \cite{chang2020mixup} & 50.1 & 61.9\\
        $\mathrm{Chang\:\textit{et al.}\;_{CVPR20}}$ \cite{chang2020weakly} & 50.9 & 63.4\\
        $\mathrm{SEAM\;_{CVPR20}}$ \cite{wang2020self} & 55.4 & 63.6\\
        $\mathrm{AdvCAM\;_{CVPR21}}$ \cite{lee2021anti} & 55.6 & 67.8\\
        $\mathrm{CPN\;_{ICCV21}}$ \cite{zhang2021complementary} & 57.4 & 68.0\\
        \hline
        TransCAM & \textbf{64.9} & \textbf{70.2}\\
        \hline
    \end{tabular}
    \caption{Comparison of pseudo labels with and without post-processing (p.p.) in terms of mIoU on PASCAL VOC 2012 \textit{train} set}
    \label{tab:pseudo}
\end{table}

\begin{table}[htbp]
    \centering
    \setlength\tabcolsep{1pt}
    \begin{tabular}{p{0.43\linewidth}p{0.21\linewidth}p{0.13\linewidth}|p{0.09\linewidth}p{0.09\linewidth}}
        \hline
        Method & Backbone & Sup. & \textit{val} & \textit{test}  \\
        \hline
        $\mathrm{PSA\;_{CVPR18}}$ \cite{ahn2018learning}  & ResNet38 & $\mathcal{I}$ & 61.7 & 63.7 \\
        $\mathrm{IRN\;_{CVPR19}}$ \cite{ahn2019weakly} & ResNet50 & $\mathcal{I}$ & 63.5 & 64.8 \\
        $\mathrm{FickleNet\;_{CVPR19}}$ \cite{lee2019ficklenet} & ResNet101 & $\mathcal{I+S}$ & 64.9 & 65.3\\
        $\mathrm{CIAN\;_{AAAI20}}$ \cite{fan2020cian} & ResNet101 & $\mathcal{I}$ & 64.3 & 65.3\\
        $\mathrm{SSDD\;_{ICCV19}}$ \cite{shimoda2019self} & ResNet38 & $\mathcal{I}$ & 64.9 & 65.5\\
        $\mathrm{SEAM\;_{CVPR20}}$ \cite{wang2020self} & ResNet38 & $\mathcal{I}$ & 64.5 & 65.7\\
        $\mathrm{Chang\:\textit{et al.}\;_{CVPR20}}$ \cite{chang2020weakly} & ResNet101 & $\mathcal{I}$ & 66.1 & 65.9\\
        $\mathrm{BES\;_{ECCV20}}$ \cite{chen2020weakly} & ResNet101 & $\mathcal{I}$ & 65.7 & 66.6\\
        $\mathrm{CONTA\;_{NIPS20}}$ \cite{zhang2020causal} & ResNet38 & $\mathcal{I}$ & 66.1 & 66.7\\
        $\mathrm{MCIS\;_{ECCV20}}$ \cite{sun2020mining} & ResNet101 & $\mathcal{I+S}$ & 66.2 & 66.9\\
        $\mathrm{ICD\;_{CVPR20}}$ \cite{fan2020learning} & ResNet101 & $\mathcal{I+S}$ & 67.8 & 68.0\\
        $\mathrm{AdvCAM\;_{CVPR21}}$ \cite{lee2021anti} & ResNet101 & $\mathcal{I}$ & 68.1 & 68.0\\
        $\mathrm{Yao\:\textit{et al.}\;_{CVPR21}}$ \cite{yao2021non} & ResNet101 & $\mathcal{I+S}$ & 68.3 & 68.5\\
        $\mathrm{CPN\;_{ICCV21}}$ \cite{zhang2021complementary} & ResNet38 & $\mathcal{I}$ & 67.8 & 68.5\\
        $\mathrm{AuxSegNet\;_{ICCV21}}$ \cite{xu2021leveraging} & ResNet38 & $\mathcal{I+S}$ & 69.0 & 68.6 \\
        \hline
        TransCAM (ours) & ResNet38 & $\mathcal{I}$ & \textbf{69.3} & \textbf{69.6}\\
        \hline
    \end{tabular}
    \caption{Comparison with state-of-the-art methods on PASCAL VOC \textit{val} and \textit{test} set in terms of mIoU (\%). The Backbone refers to the segmentation network. Sup. specifies whether image-level labels ($\mathcal{I}$) or  off-the-shelf saliency maps ($\mathcal{S}$) are used.}
    \label{tab:sota}
\end{table}

\subsection{Comparison with State-of-the-art}
We adopt PSA \cite{ahn2018learning} for post-processing following previous work \cite{wang2020self,zhang2021complementary} to improve the mIoU of our pixel-level pseudo labels. The PSA-refined pseudo labels achieve 69.14\% mIoU on the PASCAL VOC 2012 \textit{train} set. Then, the dense CRF \cite{krahenbuhl2011efficient} is employed to improve the pseudo labels obtained in the previous step. The final pseudo labels achieve 70.23\% mIoU on the PASCAL VOC 2012 \textit{train} set. Table \ref{tab:pseudo} compares the pseudo labels obtained by TransCAM (with and without post-processing) with other recent methods. For pseudo labels without post-processing, TransCAM outperforms others by a large margin; it has a smaller performance improvement after applying post-processing, possibly since it has already leveraged semantic affinities that PSA relies on for refinement. 

We train the classical segmentation model DeepLab \cite{chen2014semantic} with ResNet38 \cite{wu2019wider} backbone (pre-trained on ImageNet) on the post-processed pseudo labels in a fully-supervised manner. Table \ref{tab:categorical} shows our categorical mIoU performance on the PASCAL VOC 2012 \textit{val} set. Figure \ref{fig:qualitative} gives some qualitative semantic segmentation results on the PASCAL VOC 2012 \textit{val} set, which shows that TransCAM performs well on some relatively hard examples where multiple objects are presented in an image. Table \ref{tab:sota} compares the performance of TransCAM with previous methods. TransCAM achieves 69.3\% and 69.6\% mIoU on the PASCAL VOC 2012 \textit{val} and \textit{test} sets, respectively, which represents the state-of-the-art performance for approaches that adopt the ResNet38 segmentation backbone. Moreover, TransCAM performs 1\% better on the \textit{test} set than the previous state-of-the-art AugSegNet \cite{xu2021leveraging} method that uses additional saliency supervision. 

\section{Conclusion}
We proposed TransCAM, a transformer-CNN hybrid solution to solve limitations of class activiation maps (CAM) for weakly supervised semantic segmentation (WSSS) due to partial activation that is intrinsic to the use of CNNs. Specifically, TransCAM explicitly leverages the attention weights from the transformer branch of the Conformer to refine the CAM generated from the CNN branch. TransCAM is motivated by our observation that attention weights from shallow transformer blocks are able to capture low-level spatial feature similarities while attention weights from deep transformer blocks capture high-level semantic context. Our method is simple, efficient, and effective for generating high-quality pixel-level pseudo labels as the first transformer-based approach in WSSS. Despite its simplicity, TransCAM achieves a new state-of-the-art performance on both the PASCAL VOC 2012 validation and test sets.

{\small
\bibliographystyle{ieee_fullname}
\bibliography{egbib}

\begin{thebibliography}{10}\itemsep=-1pt

\bibitem{ahn2019weakly}
Jiwoon Ahn, Sunghyun Cho, and Suha Kwak.
\newblock Weakly supervised learning of instance segmentation with inter-pixel
  relations.
\newblock In {\em Proceedings of the IEEE/CVF conference on computer vision and
  pattern recognition}, pages 2209--2218, 2019.

\bibitem{ahn2018learning}
Jiwoon Ahn and Suha Kwak.
\newblock Learning pixel-level semantic affinity with image-level supervision
  for weakly supervised semantic segmentation.
\newblock In {\em Proceedings of the IEEE conference on computer vision and
  pattern recognition}, pages 4981--4990, 2018.

\bibitem{badrinarayanan2017segnet}
Vijay Badrinarayanan, Alex Kendall, and Roberto Cipolla.
\newblock Segnet: A deep convolutional encoder-decoder architecture for image
  segmentation.
\newblock {\em IEEE transactions on pattern analysis and machine intelligence},
  39(12):2481--2495, 2017.

\bibitem{bearman2016s}
Amy Bearman, Olga Russakovsky, Vittorio Ferrari, and Li Fei-Fei.
\newblock What’s the point: Semantic segmentation with point supervision.
\newblock In {\em European conference on computer vision}, pages 549--565.
  Springer, 2016.

\bibitem{chang2020mixup}
Yu-Ting Chang, Qiaosong Wang, Wei-Chih Hung, Robinson Piramuthu, Yi-Hsuan Tsai,
  and Ming-Hsuan Yang.
\newblock Mixup-cam: Weakly-supervised semantic segmentation via uncertainty
  regularization.
\newblock {\em arXiv preprint arXiv:2008.01201}, 2020.

\bibitem{chang2020weakly}
Yu-Ting Chang, Qiaosong Wang, Wei-Chih Hung, Robinson Piramuthu, Yi-Hsuan Tsai,
  and Ming-Hsuan Yang.
\newblock Weakly-supervised semantic segmentation via sub-category exploration.
\newblock In {\em Proceedings of the IEEE/CVF Conference on Computer Vision and
  Pattern Recognition}, pages 8991--9000, 2020.

\bibitem{chen2020weakly}
Liyi Chen, Weiwei Wu, Chenchen Fu, Xiao Han, and Yuntao Zhang.
\newblock Weakly supervised semantic segmentation with boundary exploration.
\newblock In {\em European Conference on Computer Vision}, pages 347--362.
  Springer, 2020.

\bibitem{chen2014semantic}
Liang-Chieh Chen, George Papandreou, Iasonas Kokkinos, Kevin Murphy, and Alan~L
  Yuille.
\newblock Semantic image segmentation with deep convolutional nets and fully
  connected crfs.
\newblock {\em arXiv preprint arXiv:1412.7062}, 2014.

\bibitem{chen2017deeplab}
Liang-Chieh Chen, George Papandreou, Iasonas Kokkinos, Kevin Murphy, and Alan~L
  Yuille.
\newblock Deeplab: Semantic image segmentation with deep convolutional nets,
  atrous convolution, and fully connected crfs.
\newblock {\em IEEE transactions on pattern analysis and machine intelligence},
  40(4):834--848, 2017.

\bibitem{chen2021mobile}
Yinpeng Chen, Xiyang Dai, Dongdong Chen, Mengchen Liu, Xiaoyi Dong, Lu Yuan,
  and Zicheng Liu.
\newblock Mobile-former: Bridging mobilenet and transformer.
\newblock {\em arXiv preprint arXiv:2108.05895}, 2021.

\bibitem{chen2021lctr}
Zhiwei Chen, Changan Wang, Yabiao Wang, Guannan Jiang, Yunhang Shen, Ying Tai,
  Chengjie Wang, Wei Zhang, and Liujuan Cao.
\newblock Lctr: On awakening the local continuity of transformer for weakly
  supervised object localization.
\newblock {\em arXiv preprint arXiv:2112.05291}, 2021.

\bibitem{dai2015boxsup}
Jifeng Dai, Kaiming He, and Jian Sun.
\newblock Boxsup: Exploiting bounding boxes to supervise convolutional networks
  for semantic segmentation.
\newblock In {\em Proceedings of the IEEE international conference on computer
  vision}, pages 1635--1643, 2015.

\bibitem{dai2021coatnet}
Zihang Dai, Hanxiao Liu, Quoc Le, and Mingxing Tan.
\newblock Coatnet: Marrying convolution and attention for all data sizes.
\newblock {\em Advances in Neural Information Processing Systems}, 34, 2021.

\bibitem{dosovitskiy2020image}
Alexey Dosovitskiy, Lucas Beyer, Alexander Kolesnikov, Dirk Weissenborn,
  Xiaohua Zhai, Thomas Unterthiner, Mostafa Dehghani, Matthias Minderer, Georg
  Heigold, Sylvain Gelly, et~al.
\newblock An image is worth 16x16 words: Transformers for image recognition at
  scale.
\newblock {\em arXiv preprint arXiv:2010.11929}, 2020.

\bibitem{everingham2010pascal}
Mark Everingham, Luc Van~Gool, Christopher~KI Williams, John Winn, and Andrew
  Zisserman.
\newblock The pascal visual object classes (voc) challenge.
\newblock {\em International journal of computer vision}, 88(2):303--338, 2010.

\bibitem{fan2020learning}
Junsong Fan, Zhaoxiang Zhang, Chunfeng Song, and Tieniu Tan.
\newblock Learning integral objects with intra-class discriminator for
  weakly-supervised semantic segmentation.
\newblock In {\em Proceedings of the IEEE/CVF Conference on Computer Vision and
  Pattern Recognition}, pages 4283--4292, 2020.

\bibitem{fan2020cian}
Junsong Fan, Zhaoxiang Zhang, Tieniu Tan, Chunfeng Song, and Jun Xiao.
\newblock Cian: Cross-image affinity net for weakly supervised semantic
  segmentation.
\newblock In {\em Proceedings of the AAAI Conference on Artificial
  Intelligence}, volume~34, pages 10762--10769, 2020.

\bibitem{gao2021ts}
Wei Gao, Fang Wan, Xingjia Pan, Zhiliang Peng, Qi Tian, Zhenjun Han, Bolei
  Zhou, and Qixiang Ye.
\newblock Ts-cam: Token semantic coupled attention map for weakly supervised
  object localization.
\newblock In {\em Proceedings of the IEEE/CVF International Conference on
  Computer Vision}, pages 2886--2895, 2021.

\bibitem{hariharan2011semantic}
Bharath Hariharan, Pablo Arbel{\'a}ez, Lubomir Bourdev, Subhransu Maji, and
  Jitendra Malik.
\newblock Semantic contours from inverse detectors.
\newblock In {\em 2011 international conference on computer vision}, pages
  991--998. IEEE, 2011.

\bibitem{he2016deep}
Kaiming He, Xiangyu Zhang, Shaoqing Ren, and Jian Sun.
\newblock Deep residual learning for image recognition.
\newblock In {\em Proceedings of the IEEE conference on computer vision and
  pattern recognition}, pages 770--778, 2016.

\bibitem{hou2018self}
Qibin Hou, PengTao Jiang, Yunchao Wei, and Ming-Ming Cheng.
\newblock Self-erasing network for integral object attention.
\newblock {\em Advances in Neural Information Processing Systems}, 31, 2018.

\bibitem{hwang2016self}
Sangheum Hwang and Hyo-Eun Kim.
\newblock Self-transfer learning for weakly supervised lesion localization.
\newblock In {\em International conference on medical image computing and
  computer-assisted intervention}, pages 239--246. Springer, 2016.

\bibitem{jo2021puzzle}
Sanghyun Jo and In-Jae Yu.
\newblock Puzzle-cam: Improved localization via matching partial and full
  features.
\newblock In {\em 2021 IEEE International Conference on Image Processing
  (ICIP)}, pages 639--643. IEEE, 2021.

\bibitem{khoreva2017simple}
Anna Khoreva, Rodrigo Benenson, Jan Hosang, Matthias Hein, and Bernt Schiele.
\newblock Simple does it: Weakly supervised instance and semantic segmentation.
\newblock In {\em Proceedings of the IEEE conference on computer vision and
  pattern recognition}, pages 876--885, 2017.

\bibitem{kolesnikov2016seed}
Alexander Kolesnikov and Christoph~H Lampert.
\newblock Seed, expand and constrain: Three principles for weakly-supervised
  image segmentation.
\newblock In {\em European conference on computer vision}, pages 695--711.
  Springer, 2016.

\bibitem{krahenbuhl2011efficient}
Philipp Krähenbühl and Vladlen Koltun.
\newblock Efficient inference in fully connected crfs with gaussian edge
  potentials.
\newblock {\em Advances in neural information processing systems}, 24, 2011.

\bibitem{lee2019ficklenet}
Jungbeom Lee, Eunji Kim, Sungmin Lee, Jangho Lee, and Sungroh Yoon.
\newblock Ficklenet: Weakly and semi-supervised semantic image segmentation
  using stochastic inference.
\newblock In {\em Proceedings of the IEEE/CVF Conference on Computer Vision and
  Pattern Recognition}, pages 5267--5276, 2019.

\bibitem{lee2021anti}
Jungbeom Lee, Eunji Kim, and Sungroh Yoon.
\newblock Anti-adversarially manipulated attributions for weakly and
  semi-supervised semantic segmentation.
\newblock In {\em Proceedings of the IEEE/CVF Conference on Computer Vision and
  Pattern Recognition}, pages 4071--4080, 2021.

\bibitem{li2018tell}
Kunpeng Li, Ziyan Wu, Kuan-Chuan Peng, Jan Ernst, and Yun Fu.
\newblock Tell me where to look: Guided attention inference network.
\newblock In {\em Proceedings of the IEEE Conference on Computer Vision and
  Pattern Recognition}, pages 9215--9223, 2018.

\bibitem{lin2016scribblesup}
Di Lin, Jifeng Dai, Jiaya Jia, Kaiming He, and Jian Sun.
\newblock Scribblesup: Scribble-supervised convolutional networks for semantic
  segmentation.
\newblock In {\em Proceedings of the IEEE conference on computer vision and
  pattern recognition}, pages 3159--3167, 2016.

\bibitem{long2015fully}
Jonathan Long, Evan Shelhamer, and Trevor Darrell.
\newblock Fully convolutional networks for semantic segmentation.
\newblock In {\em Proceedings of the IEEE conference on computer vision and
  pattern recognition}, pages 3431--3440, 2015.

\bibitem{loshchilov2017decoupled}
Ilya Loshchilov and Frank Hutter.
\newblock Decoupled weight decay regularization.
\newblock {\em arXiv preprint arXiv:1711.05101}, 2017.

\bibitem{peng2021conformer}
Zhiliang Peng, Wei Huang, Shanzhi Gu, Lingxi Xie, Yaowei Wang, Jianbin Jiao,
  and Qixiang Ye.
\newblock Conformer: Local features coupling global representations for visual
  recognition.
\newblock In {\em Proceedings of the IEEE/CVF International Conference on
  Computer Vision}, pages 367--376, 2021.

\bibitem{russakovsky2015imagenet}
Olga Russakovsky, Jia Deng, Hao Su, Jonathan Krause, Sanjeev Satheesh, Sean Ma,
  Zhiheng Huang, Andrej Karpathy, Aditya Khosla, Michael Bernstein, et~al.
\newblock Imagenet large scale visual recognition challenge.
\newblock {\em International journal of computer vision}, 115(3):211--252,
  2015.

\bibitem{shimoda2019self}
Wataru Shimoda and Keiji Yanai.
\newblock Self-supervised difference detection for weakly-supervised semantic
  segmentation.
\newblock In {\em Proceedings of the IEEE/CVF International Conference on
  Computer Vision}, pages 5208--5217, 2019.

\bibitem{simonyan2014very}
Karen Simonyan and Andrew Zisserman.
\newblock Very deep convolutional networks for large-scale image recognition.
\newblock {\em arXiv preprint arXiv:1409.1556}, 2014.

\bibitem{sun2020mining}
Guolei Sun, Wenguan Wang, Jifeng Dai, and Luc Van~Gool.
\newblock Mining cross-image semantics for weakly supervised semantic
  segmentation.
\newblock In {\em European conference on computer vision}, pages 347--365.
  Springer, 2020.

\bibitem{touvron2021training}
Hugo Touvron, Matthieu Cord, Matthijs Douze, Francisco Massa, Alexandre
  Sablayrolles, and Herv{\'e} J{\'e}gou.
\newblock Training data-efficient image transformers \& distillation through
  attention.
\newblock In {\em International Conference on Machine Learning}, pages
  10347--10357. PMLR, 2021.

\bibitem{vaswani2017attention}
Ashish Vaswani, Noam Shazeer, Niki Parmar, Jakob Uszkoreit, Llion Jones,
  Aidan~N Gomez, {\L}ukasz Kaiser, and Illia Polosukhin.
\newblock Attention is all you need.
\newblock {\em Advances in neural information processing systems}, 30, 2017.

\bibitem{wang2018non}
Xiaolong Wang, Ross Girshick, Abhinav Gupta, and Kaiming He.
\newblock Non-local neural networks.
\newblock In {\em Proceedings of the IEEE conference on computer vision and
  pattern recognition}, pages 7794--7803, 2018.

\bibitem{wang2020self}
Yude Wang, Jie Zhang, Meina Kan, Shiguang Shan, and Xilin Chen.
\newblock Self-supervised equivariant attention mechanism for weakly supervised
  semantic segmentation.
\newblock In {\em Proceedings of the IEEE/CVF Conference on Computer Vision and
  Pattern Recognition}, pages 12275--12284, 2020.

\bibitem{wei2017object}
Yunchao Wei, Jiashi Feng, Xiaodan Liang, Ming-Ming Cheng, Yao Zhao, and
  Shuicheng Yan.
\newblock Object region mining with adversarial erasing: A simple
  classification to semantic segmentation approach.
\newblock In {\em Proceedings of the IEEE conference on computer vision and
  pattern recognition}, pages 1568--1576, 2017.

\bibitem{wu2021cvt}
Haiping Wu, Bin Xiao, Noel Codella, Mengchen Liu, Xiyang Dai, Lu Yuan, and Lei
  Zhang.
\newblock Cvt: Introducing convolutions to vision transformers.
\newblock In {\em Proceedings of the IEEE/CVF International Conference on
  Computer Vision}, pages 22--31, 2021.

\bibitem{wu2021embedded}
Tong Wu, Junshi Huang, Guangyu Gao, Xiaoming Wei, Xiaolin Wei, Xuan Luo, and
  Chi~Harold Liu.
\newblock Embedded discriminative attention mechanism for weakly supervised
  semantic segmentation.
\newblock In {\em Proceedings of the IEEE/CVF Conference on Computer Vision and
  Pattern Recognition}, pages 16765--16774, 2021.

\bibitem{wu2019wider}
Zifeng Wu, Chunhua Shen, and Anton Van Den~Hengel.
\newblock Wider or deeper: Revisiting the resnet model for visual recognition.
\newblock {\em Pattern Recognition}, 90:119--133, 2019.

\bibitem{xu2021leveraging}
Lian Xu, Wanli Ouyang, Mohammed Bennamoun, Farid Boussaid, Ferdous Sohel, and
  Dan Xu.
\newblock Leveraging auxiliary tasks with affinity learning for weakly
  supervised semantic segmentation.
\newblock In {\em Proceedings of the IEEE/CVF International Conference on
  Computer Vision}, pages 6984--6993, 2021.

\bibitem{xu2021vitae}
Yufei Xu, Qiming Zhang, Jing Zhang, and Dacheng Tao.
\newblock Vitae: Vision transformer advanced by exploring intrinsic inductive
  bias.
\newblock {\em Advances in Neural Information Processing Systems}, 34, 2021.

\bibitem{yao2021non}
Yazhou Yao, Tao Chen, Guo-Sen Xie, Chuanyi Zhang, Fumin Shen, Qi Wu, Zhenmin
  Tang, and Jian Zhang.
\newblock Non-salient region object mining for weakly supervised semantic
  segmentation.
\newblock In {\em Proceedings of the IEEE/CVF Conference on Computer Vision and
  Pattern Recognition}, pages 2623--2632, 2021.

\bibitem{zhang2020causal}
Dong Zhang, Hanwang Zhang, Jinhui Tang, Xian-Sheng Hua, and Qianru Sun.
\newblock Causal intervention for weakly-supervised semantic segmentation.
\newblock {\em Advances in Neural Information Processing Systems}, 33:655--666,
  2020.

\bibitem{zhang2021complementary}
Fei Zhang, Chaochen Gu, Chenyue Zhang, and Yuchao Dai.
\newblock Complementary patch for weakly supervised semantic segmentation.
\newblock In {\em Proceedings of the IEEE/CVF International Conference on
  Computer Vision}, pages 7242--7251, 2021.

\bibitem{zhou2016learning}
Bolei Zhou, Aditya Khosla, Agata Lapedriza, Aude Oliva, and Antonio Torralba.
\newblock Learning deep features for discriminative localization.
\newblock In {\em Proceedings of the IEEE conference on computer vision and
  pattern recognition}, pages 2921--2929, 2016.

\end{thebibliography}
}

\clearpage

\setcounter{section}{0}
\renewcommand\thesection{\Alph{section}}

\section{Attention weights coupling approaches}
\begin{figure*}[hb]
    \centering
    \begin{tabular}{m{0.08\linewidth}m{0.72\linewidth}}
        \hline
        \\
        \small{ClsAttn} & \includegraphics[width=\linewidth]{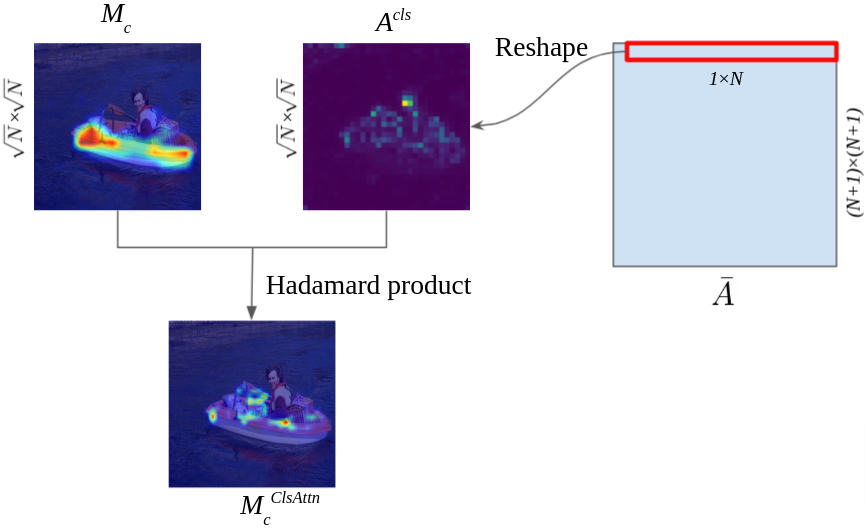}\\
        \\
        \hline
        \\
        \small{AttnAgg} & \includegraphics[width=\linewidth]{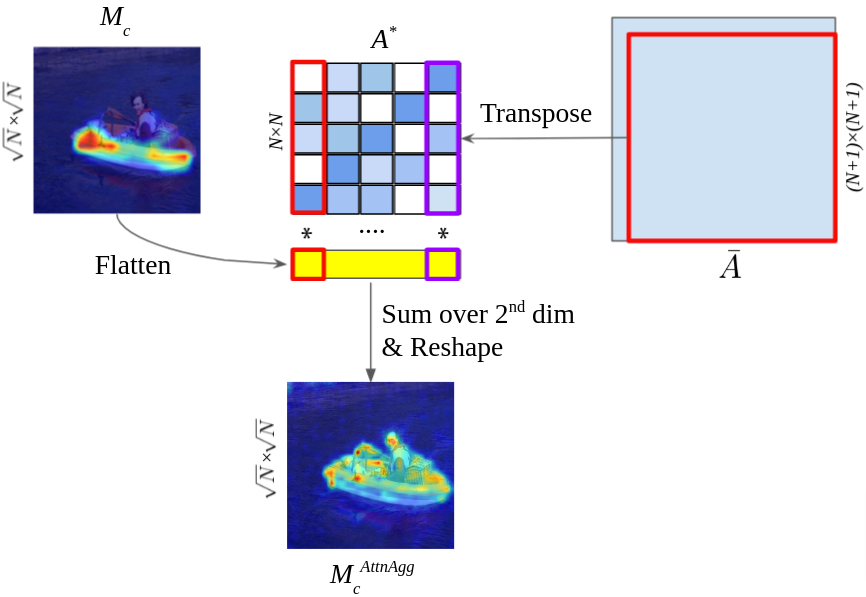}\\
        \\
        \hline
    \end{tabular}
    \caption{\textit{ClsAttn} couples CAM with attention weights that corresponds to the class token via Hadamard product; \textit{AttnAgg} aggregates the patch attention weights based on the corresponding activation score from CAM.}
    \label{fig:app_method}
\end{figure*}

This section provides additional discussion and compares TransCAM with two approaches to explicitly couple CAM with attention weights as mentioned in Section 4.3: \textit{ClsAttn} and \textit{AttnAgg}. \textit{ClsAttn} employs the attention weights for the class token while \textit{AttnAgg} aggregates the attention weights based on the activation scores from CAM. Figure \ref{fig:app_method} illustrates these two attention weight coupling approaches with the detailed calculation provided below: 

\subsection{ClsAttn}
We employ the attention weights $A^{cls}$ that attribute to the class token:
\begin{equation}
    A^{cls} = (\bar{A}_{i,j})_{i=1, 2\leq j \leq N+1}
\end{equation}
Recall that $\bar{A}$ represents the average of the self-attention weights across all blocks and all heads, and $\bar{A}_{i,j}$ denotes on average how much the $j$-th input token embedding contributes to the $i$-th output token embedding in all layers. $A^{cls} \in \mathbb{R}^{1\times N}$ highlights the crucial pixels for image classification (mainly highlights foreground objects), however, it is class-agnostic (indistinguishable for different target classes). We fuse it with the CAM $M_c$ via Hadamard product to make it aware of object categories:
\begin{equation}
  M_c^{ClsAttn} = \mathrm{\Gamma}^{\sqrt{N}\times \sqrt{N}}(A_{cls}) \odot M_c
\end{equation}
where $M_c^{ClsAttn}$ is the CAM coupled with attention weights by $ClsAttn$.

\subsection{AttnAgg}
We assume that $(A^*)^{\intercal} = [A^*_1;A^*_2;...;A^*_N]$, where $A^*_i \in \mathbb{R}^{N\times 1}$ denotes the attention weights (excluding class token weights) that attribute to the $i$-th token, highlighting the pixels correlated to the $i$-th patch. We aggregate $A^*_i$ based on the activation score of the $i$-th patch, which is determined by the value at the $i$-th location of the CAM $M_c$:
\begin{equation}
  M_c^{AttnAgg} = \mathrm{\Gamma}^{\sqrt{N}\times\sqrt{N}}(\sum_i(\mathrm{\Gamma}^{N}M_c)_i A^*_i)
\end{equation}
Compared with the class-agnostic attention weights $A^*$, $M_c^{AttnAgg}$ is aware of object categories and facilities better object localization ability.

Figure \ref{fig:abl_ac} shows the qualitative comparison of the activation maps generated by \textit{ClsAttn}, \textit{AttnAgg}, and TransCAM. \textit{ClsAttn} usually leads to under-activated CAMs due to possible extreme values at local areas. \textit{AttnAgg} leads to more extensive object coverage but at the risk of highlighting irrelevant objects since it relies on the class-agnostic attention weights for localization to a large extent. The result verifies TransCAM to be the most proper attention weights coupling approach for the WSSS task.

\section{Performance with Different Conformer Architectures}
In this section, we evaluate the initial pseudo label performance generated by TransCAM with different ImageNet\cite{russakovsky2015imagenet} pre-trained Conformer architectures:\\ Conformer-Ti, Conformer-S, and Conformer-B. These architectures mainly differ in terms of CNN and transformer embedding dimensions, and the number of heads for the multi-head self-attention modules. Table \ref{tab:abl_arch} shows these architectures, the number of parameters, the ImageNet top-1 validation accuracy, and their performance on PASCAL VOC \textit{train} set. Conformer-S outperforms the smaller size Conformer-Ti by 6.01\% mIoU. However, Conformer-B leads to a slight drop in mIoU (0.47\%), although it has more parameters and higher pretraining accuracy. Our result shows that Conformer-S is the best architecture for TransCAM, considering the model size and mIoU performance. 

\begin{table}[htbp]
    \centering
    \begin{tabular}{p{0.25\linewidth}p{0.20\linewidth}p{0.16\linewidth}|@{\hskip5pt}l}
        \hline
        architecture & params (M) & acc. (\%) &  mIoU (\%)\\
        \hline
        Conformer-Ti & 23.5 & 81.3 &  58.84\\
        Conformer-S & 37.7 & 83.4 & 64.85\\
        Conformer-B & 83.3 & 84.1 & 64.38\\
        \hline
    \end{tabular}
    \caption{Comparison of different ImageNet-pre-trained Conformer architectures along with their number of parameters and pretraining accuracy.}
    \label{tab:abl_arch}
\end{table}

\section{Additional Semantic Segmentation Results}
This section includes additional semantic segmentation results that we did not have space to include in the main paper. Table \ref{tab:cate_test} shows the categorical semantic segmentation performance on PASCAL VOC 2012 \textit{test} set. In 13 out of 21 classes, TransCAM outperforms other compared methods.

\begin{table*}[t!]
\centering
\resizebox{.95\textwidth}{!}{
\setlength\tabcolsep{1.5pt}
\begin{tabular}{l|ccccccccccccccccccccc|c}
\hline
method & bkg & aero & bike & bird & boat & bottle & bus & car & cat & chair & cow & table & dog & horse & mbk & person & plant & sheep & sofa & train & tv & mIoU\\
\hline
SEC \cite{kolesnikov2016seed} & 83.5 & 56.4 & 28.5 & 64.1 & 23.6 & 46.5 & 70.6 & 58.5 & 71.3 & 23.2 & 54.0 & 28.0 & 68.1 & 62.1 & 70.0 & 55.0 & 38.4 & 58.0 & 39.9 & 38.4 & 48.3 & 51.7\\
PSA \cite{ahn2018learning} & 89.1 & 70.6 & 31.6 & 77.2 & 42.2 & \textbf{68.9} & 79.1 & 66.5 & 74.9 & 29.6 & 68.7 & 56.1 & 82.1 & 64.8 & 78.6 & 73.5 & \textbf{50.8} & 70.7 & 47.7 & 63.9 & \textbf{51.1} & 63.7\\
SSDD \cite{shimoda2019self} & 89.5 & 71.8 & 31.4 & 79.3 & 47.3 & 64.2 & 79.9 & 74.6 & 84.9 & \textbf{30.8} & 73.5 & \textbf{58.2} & \textbf{82.7} & 73.4 & 76.4 & 69.9 & 37.4 & 80.5 & 54.5 & 65.7 & 50.3 & 65.5\\
BES\cite{chen2020weakly} & 89.0 & 72.7 & 30.4 & 84.6 & 47.5 & 63.0 & \textbf{86.8} & 80.7 & \textbf{85.2} & 30.1 & 76.5 & 56.4 & 81.8 & 79.9 & 77.0 & 67.8 & 48.6 & 82.3 & \textbf{57.2} & 54.0 & 46.7 & 66.6\\
\hline
\textbf{Ours} & \textbf{91.2} & \textbf{82.2} & \textbf{36.2} & \textbf{89.8} & \textbf{57.8} & 65.1 & 85.7 & \textbf{81.5} & 84.2 & 28.3 & \textbf{77.4} & 55.9 & 82.1 & \textbf{80.0} & \textbf{78.9} & \textbf{76.5} & 48.0 & \textbf{84.7} & \textbf{57.2} & \textbf{72.7} & 45.8 & \textbf{69.6}\\
\hline
\end{tabular}
}
\caption{Categorical semantic segmentation performance on PASCAL VOC 2012 \textit{test} set.}
\label{tab:cate_test}
\end{table*}

\begin{figure*}[t]
\centering
\resizebox{0.78\textwidth}{!}{
    \bgroup
    \def\arraystretch{0.3}
    \setlength\tabcolsep{1.2pt}
    \begin{tabular}{p{0.16\textwidth}p{0.16\textwidth}p{0.16\textwidth}p{0.16\textwidth}p{0.16\textwidth}p{0.17\textwidth}}
        \multicolumn{6}{c}{\includegraphics[width=\linewidth]{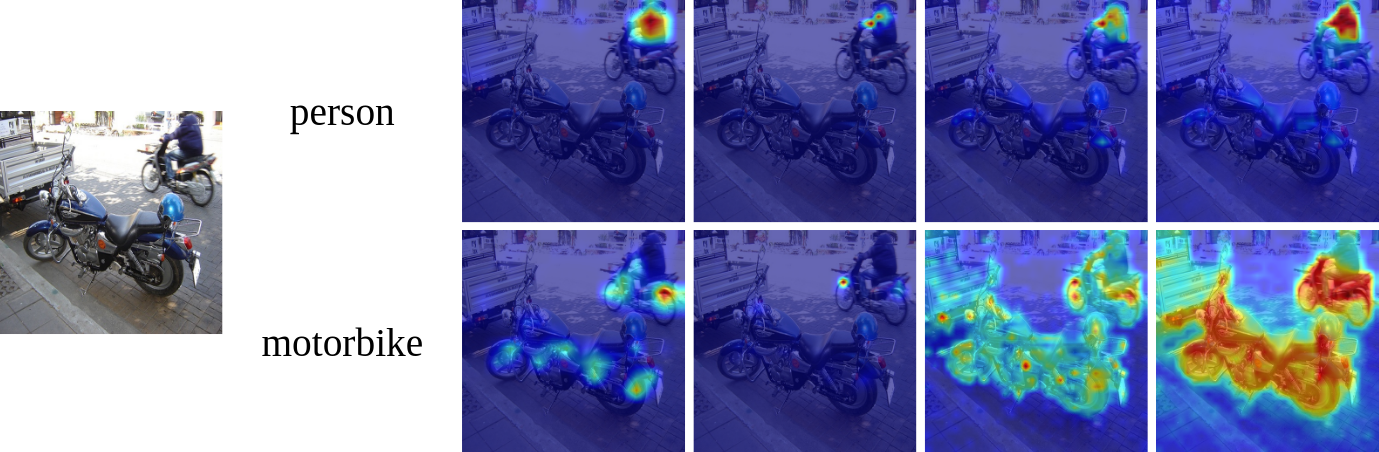}}
        \\
        \\
        \multicolumn{6}{c}{\includegraphics[width=\linewidth]{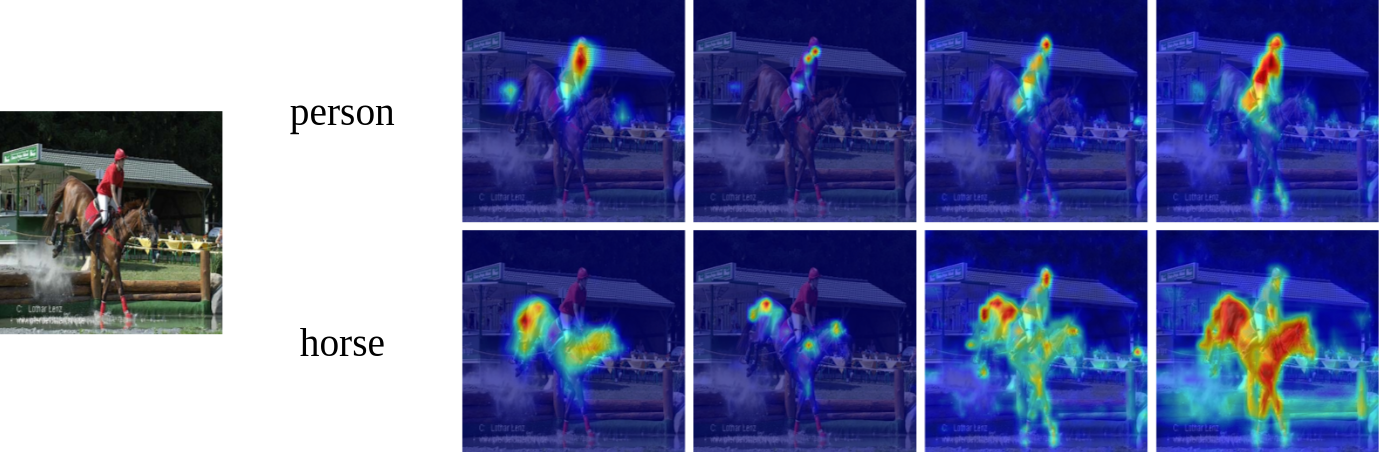}}
        \\
        \\
        \multicolumn{6}{c}{\includegraphics[width=\linewidth]{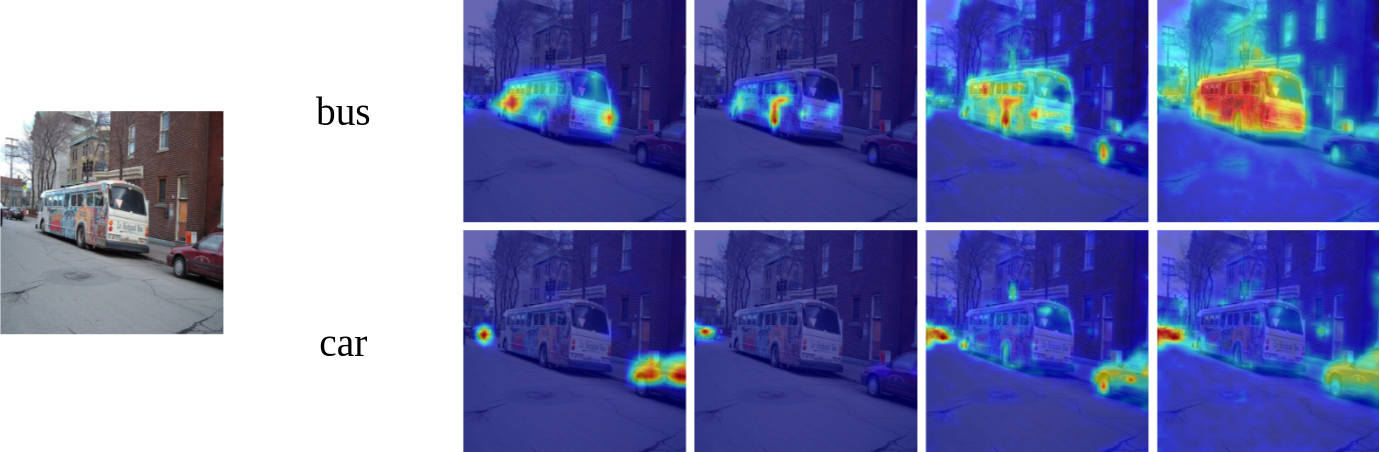}}
        \\
        \\
        \multicolumn{6}{c}{\includegraphics[width=\linewidth]{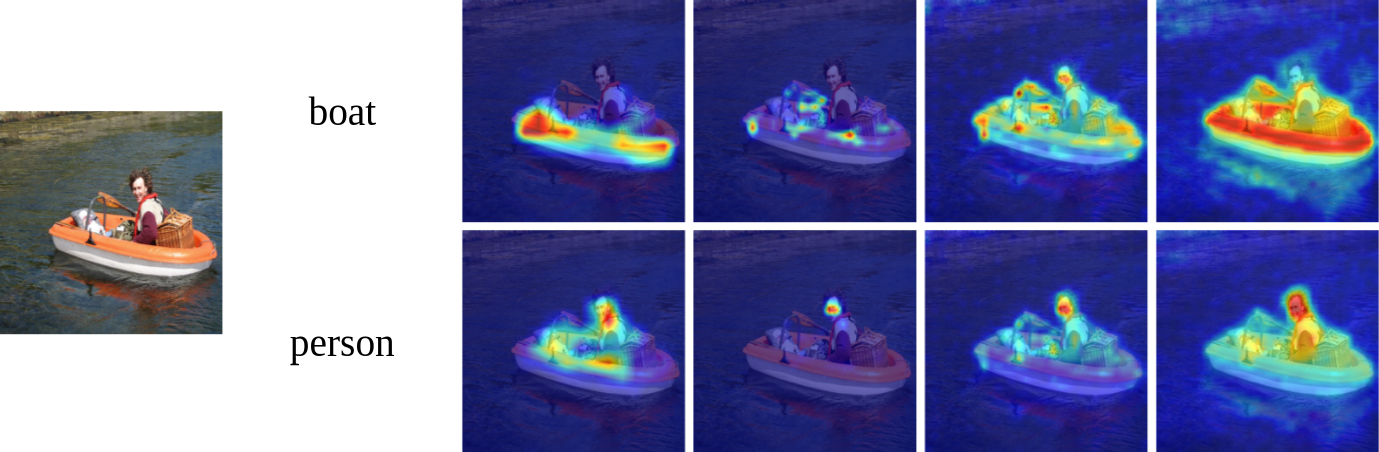}}
        \\
        \\
        \centering (a) image & \centering (b) label & \centering (c) baseline & \centering (d) ClsAttn & \centering (e) AttnAgg & \centering (f) TransCAM\\
    \end{tabular}
    \egroup
    }
    \caption{Qualitative comparison of different attention coupling methods. (a) input image; (b) the target class label; (c) Conformer-S-CAM (baseline); (d) ClsAttn; (e) AttnAgg; (f) TransCAM (ours)}
    \label{fig:abl_ac}
\end{figure*}

\end{document}